\documentclass{article}

\usepackage{arxiv}
\usepackage{amsmath,amssymb,amsfonts}
\usepackage{bm} 
\usepackage{url}
\usepackage{hyperref}
\usepackage[english]{babel}
\usepackage{graphicx}
\usepackage{fixmetodonotes}
\usepackage[table]{xcolor}
\usepackage{tikz}
\usepackage{caption} 
\usepackage{wrapfig} 
\usepackage[square,numbers]{natbib}

\renewcommand{\vec}[1]{\mathbf{#1}}
\newcommand{\matr}[1]{\mathbf{#1}}
\newcommand{\wmatr}[1]{\matr{W}^{\text{#1}}}
\newcommand{\Whidden}{\wmatr{h}}
\newcommand{\Wout}{\wmatr{out}}
\newcommand{\Win}{\wmatr{in}}
\newcommand{\Winf}{\mathsf{Win}}

\newcommand{\vt}[1]{\vec{#1}_t}
\newcommand{\norm}[1]{\left\|#1\right\|}
\newcommand{\bigO}[1]{\mathcal{O}\left(#1\right)}
\newcommand{\floor}[1]{\left\lfloor #1 \right\rfloor}

\DeclareMathOperator*{\argmin}{arg\,min}

\graphicspath{{figures/}}

\newlength{\fontsizelength}

\newcommand{\ExtendedESN}{
  \usetikzlibrary{shapes.geometric}

  \newcommand{\lw}{\linewidth}

  \begin{center}
  \begin{tikzpicture}[scale=1]

    \node[inner sep=0pt] (input) at (0,0) {
      \includegraphics[width=0.1\linewidth]{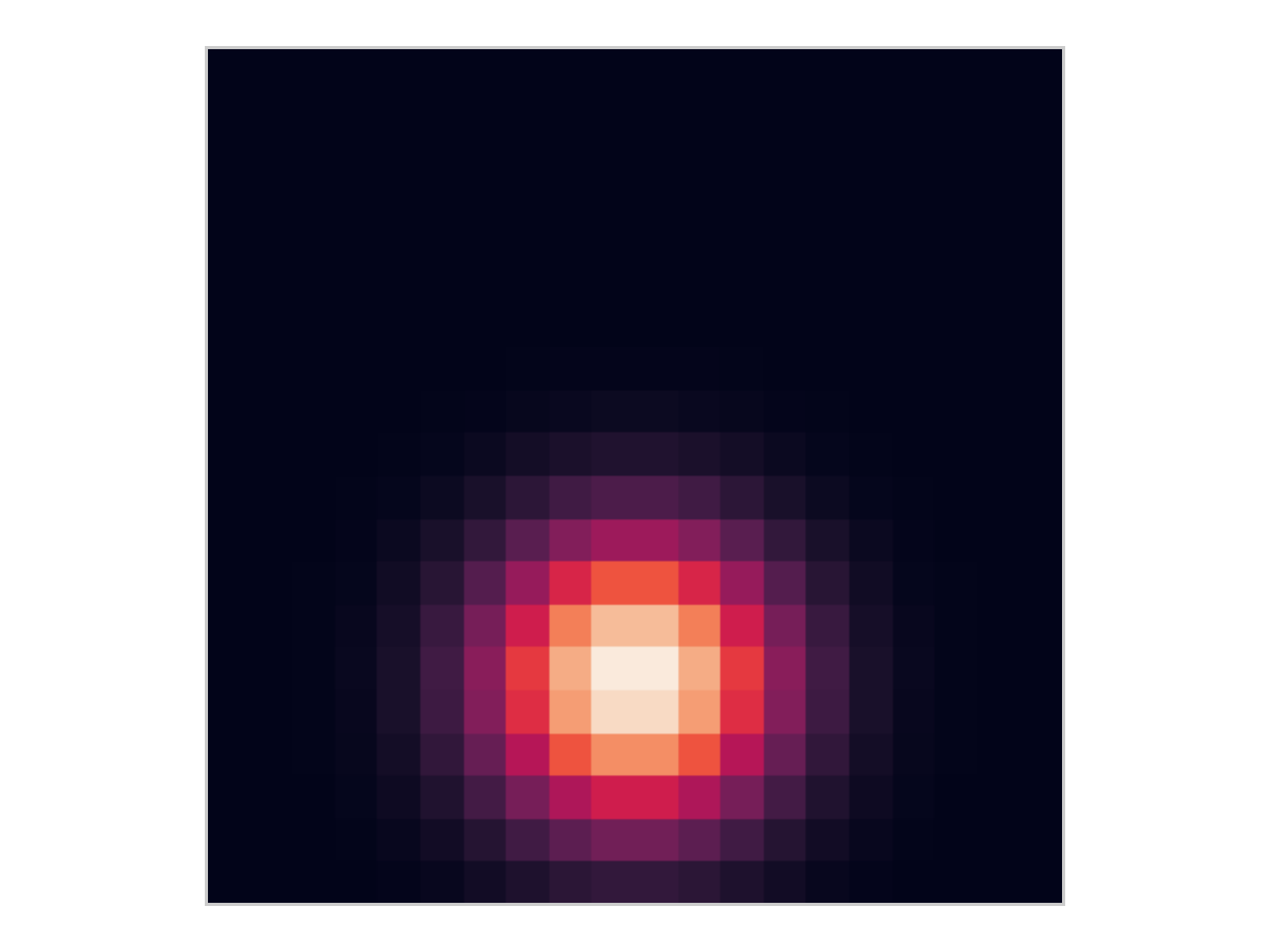}};
    \node (input_name) at (-0.1\lw, 0) {Input $u_i$};
    \node[draw] (Win) at (0.2\lw, 0.24\lw) {$\mathbf{Win}(u_i)$};

    \node[inner sep=0pt] (pixel)    at (0.2\lw, 0.16\lw) {
      \includegraphics[width=0.1\linewidth]{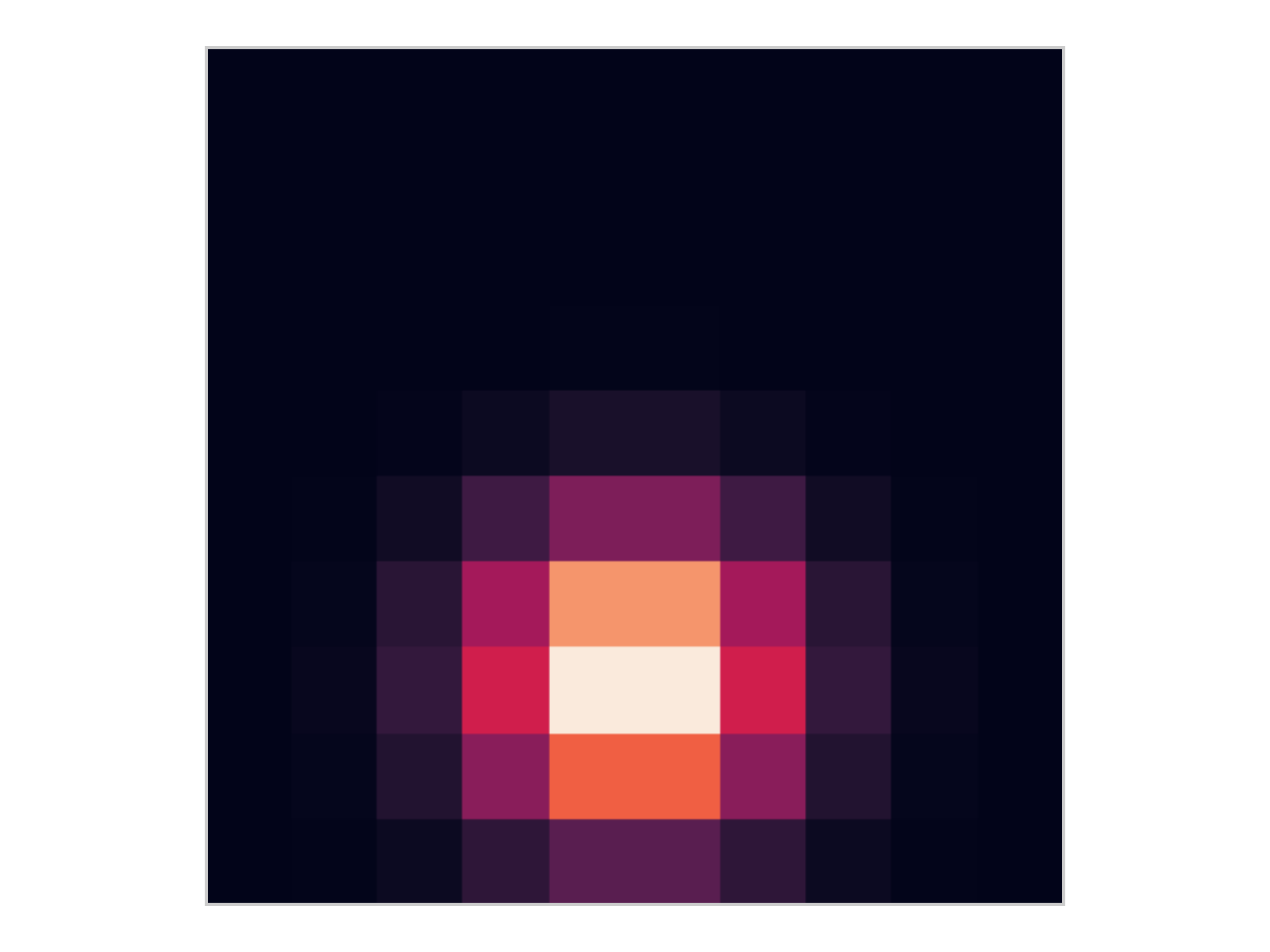}};
    \node (pixel_name) at (0.28\lw, 0.16\lw) {Pixel};

    \node[inner sep=0pt] (conv)     at (0.2\lw, 0.08\lw) {
      \includegraphics[width=0.1\linewidth]{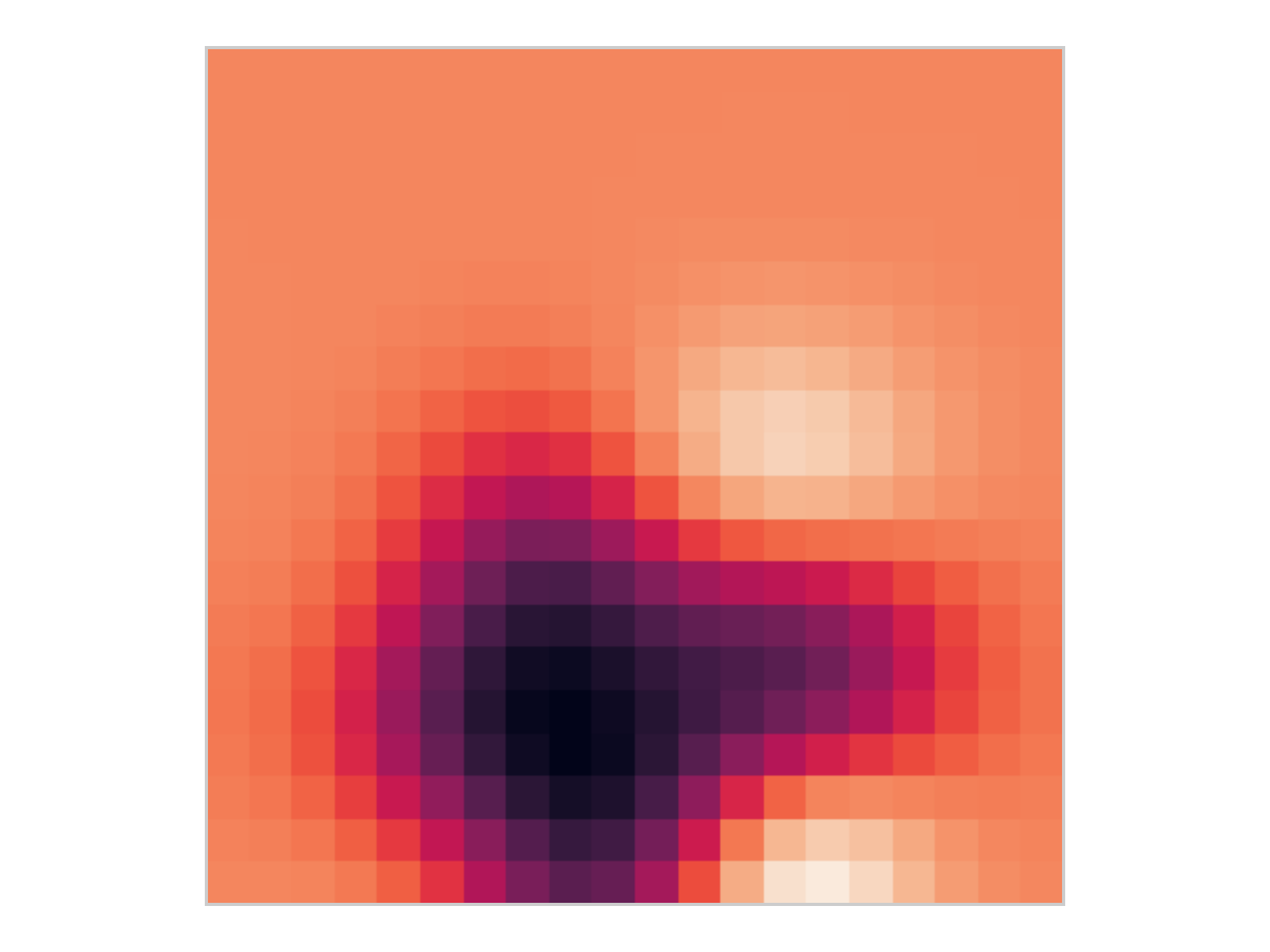}};
    \node (conv_name) at (0.28\lw, 0.08\lw) {Conv.};

    \node[inner sep=0pt] (dct)      at (0.2\lw, 0) {
      \includegraphics[width=0.1\linewidth]{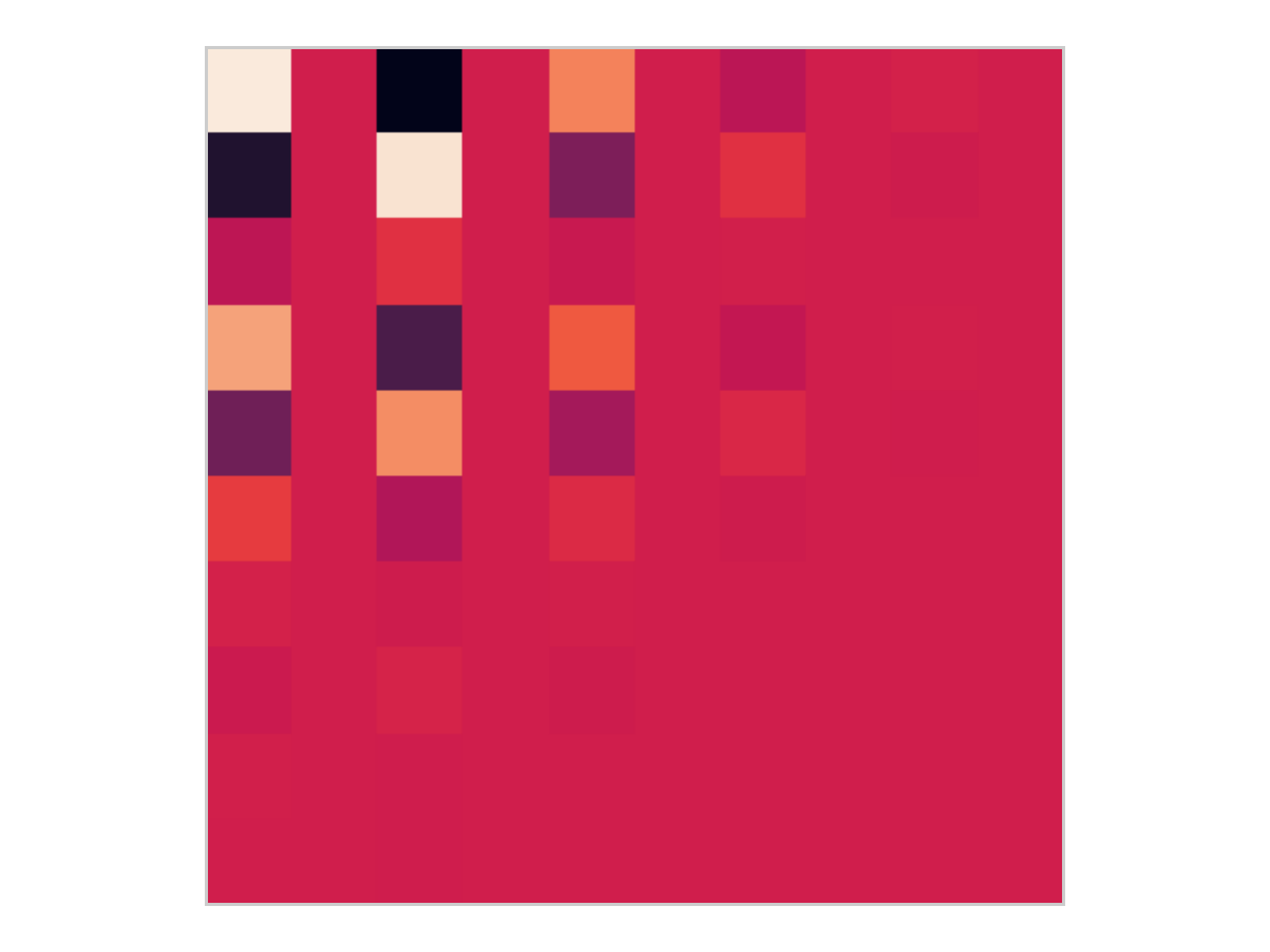}};
    \node (dct_name) at (0.28\lw, 0) {DCT};

    \node[inner sep=0pt] (gradient) at (0.2\lw, -0.11\lw) {
      \includegraphics[width=0.2\linewidth]{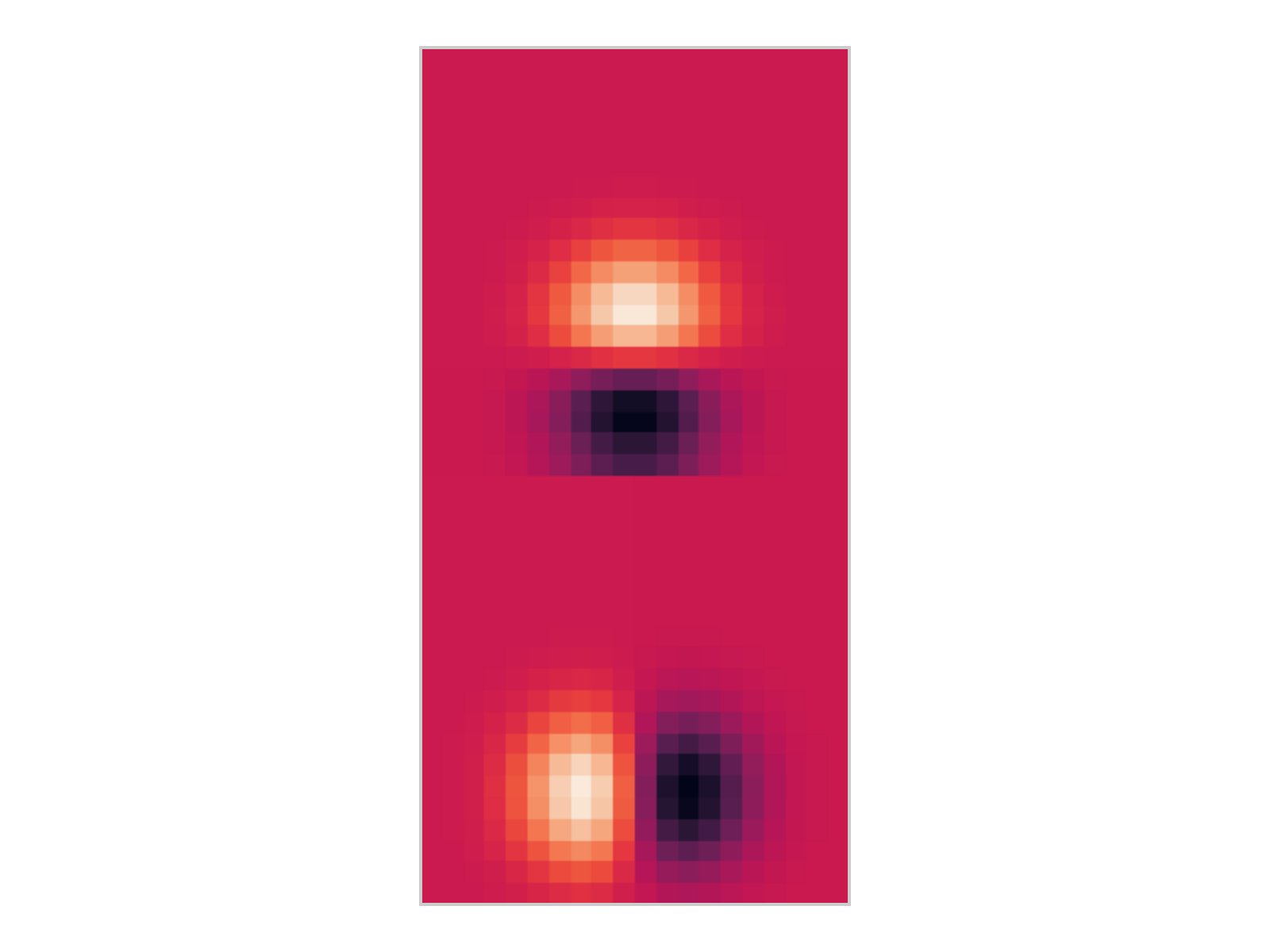}};
    \node (grad_name) at (0.28\lw, -0.11\lw) {Grad.};

    \node[inner sep=0pt] (random)   at (0.2\lw, -0.22\lw) {
      \includegraphics[width=0.1\linewidth]{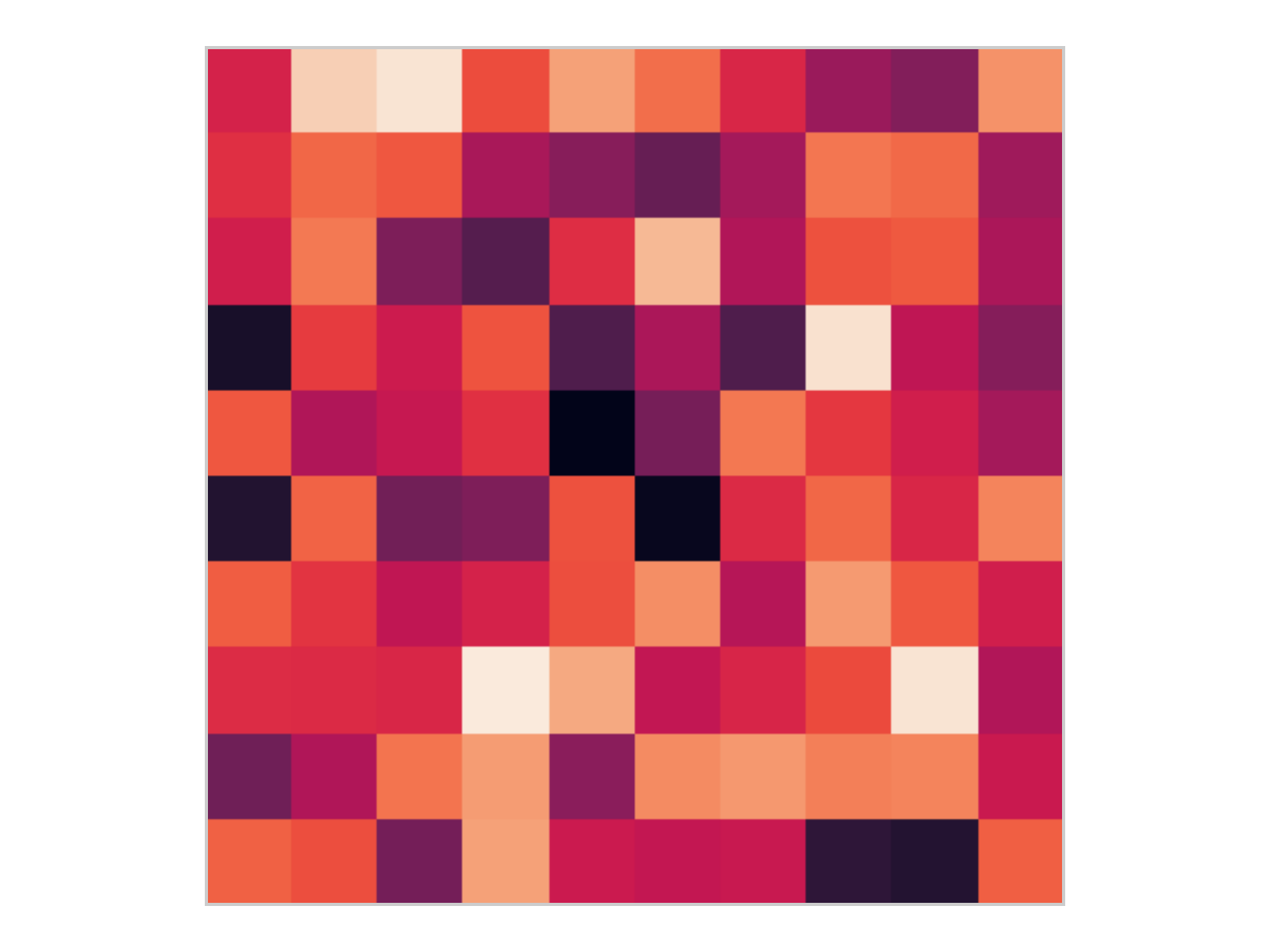}};
    \node (random_name) at (0.28\lw, -0.22\lw) {R.W.};

    \path[draw, ->, thick] (input) -- (pixel) node {};
    \path[draw, ->, thick] (input) -- (conv) node {};
    \path[draw, ->, thick] (input) -- (dct) node {};
    \path[draw, ->, thick, shorten >= -0.06\lw] (input) -- (gradient) node {};
    \path[draw, ->, thick, shorten >= 0.01\lw, shorten <= 0.01\lw] (input) -- (random) node {};
    \node (plus) at (0.34\lw, 0.24\lw) {$+$};

    \newcommand{\xstate}{0.46\lw}
    \node (state) at (\xstate, 0) {};
    \node (state_top) at (\xstate+0.015\lw, 0.28\lw) {};
    \draw[draw=black, fill=gray, thick]  (\xstate, -0.32\lw) rectangle ++(0.03\lw, 0.6\lw);
    \node[fill=white, draw=black] (W) at (\xstate+0.015\lw, 0.24\lw) {$\mathbf{W}x_i$};

    \path[draw, ->, ultra thick, shorten <= 0.01\lw] (dct_name) -- (state) node {};
    \path[draw, ultra thick] (0.33\lw, 0.19\lw) -- (0.33\lw, -0.25\lw) node {};
    \path[draw, ->, ultra thick] (state_top) edge[in=130, out=50, loop, distance=0.1\lw] (state_top) node {};

    \node[inner sep=0pt] (output)    at (0.65\lw, 0) {
      \includegraphics[width=0.1\linewidth]{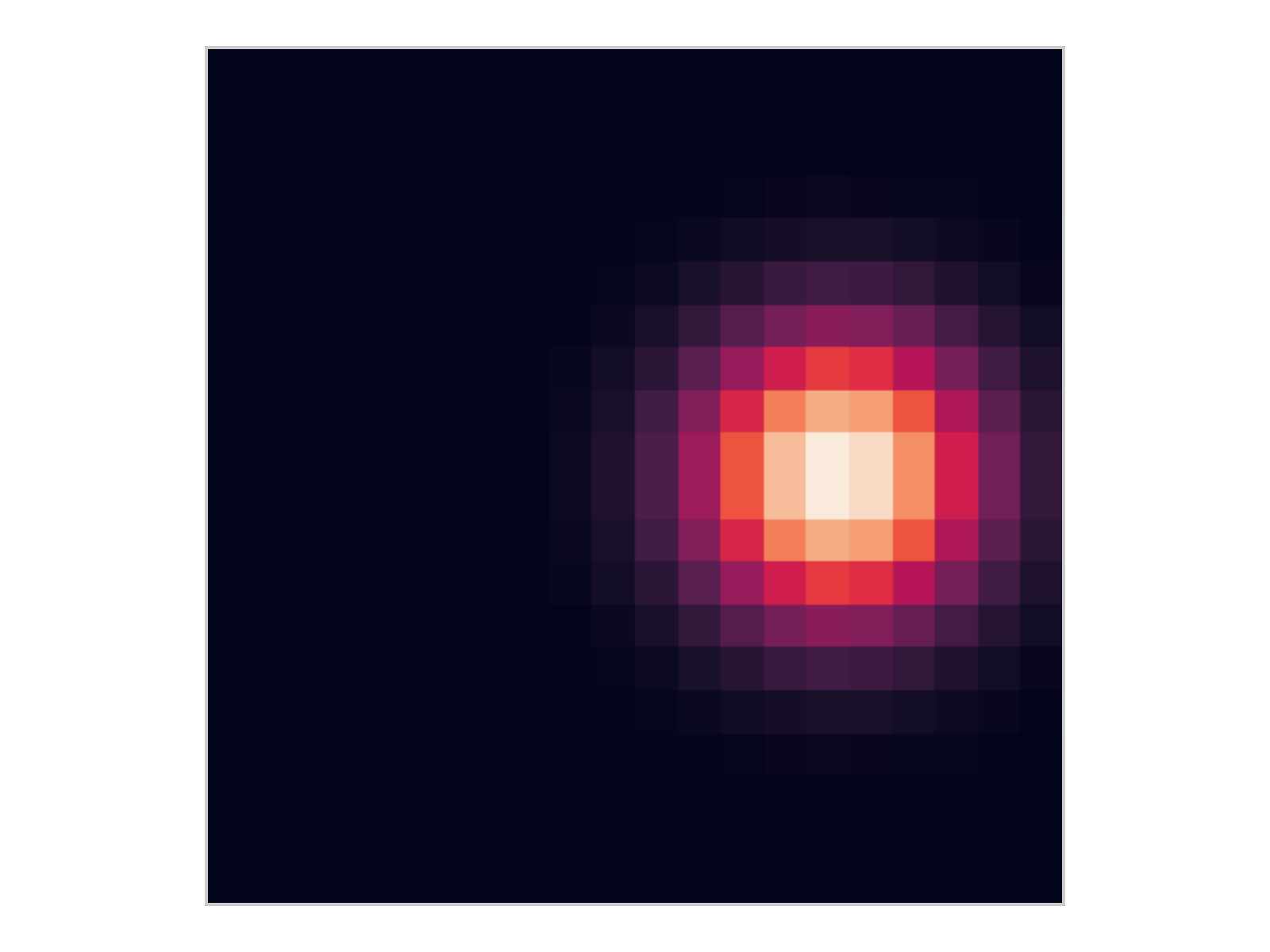}};
    \node (output_name) at (0.74\lw, 0) {Output};
    \path[draw, ->, ultra thick, shorten <= 0.032\lw] (state) -- (output) node {};
    \node[draw] (Wout) at (0.65\lw, 0.24\lw) {$\Wout y_i$};

  \end{tikzpicture}
  \end{center}
}

\begin{document}

\title{Adaptive Anomaly Detection in Chaotic Time Series
with a Spatially Aware Echo State Network}

\author{Niklas Heim \\
  Artificial Intelligence Center \\
  Czech Technical University\\
  Prague, Czech Republic\\
  \texttt{niklas.heim@aic.fel.cvut.cz} \\
  \And
  James E. Avery \\
  Niels Bohr Institute,\\
  University of Copenhagen\\
  Copenhagen, Denmark \\
  \texttt{avery@nbi.dk}
}

\maketitle

\begin{abstract}
  This work builds an automated anomaly detection method for
  chaotic time series, and more concretely for turbulent, high-dimensional,
  ocean simulations.\\
  We solve this task by extending the \emph{Echo State Network}
  \cite{jaeger2001} by spatially aware input maps, such as convolutions, gradients, cosine transforms,
  et cetera, as well as a spatially aware loss function.  The
  \emph{spatial ESN} is used to create predictions which reduce the detection problem
  to thresholding of the prediction error.\\
  We benchmark our detection framework on different tasks of
  increasing difficulty to show the generality of the framework before applying
  it to raw climate model output in the region of the Japanese ocean current
  Kuroshio, which exhibits a bimodality that is not easily detected by the
  naked eye. The code is available as an open source Python package,
  {\em Torsk}, available at \url{https://github.com/nmheim/torsk},
  where we also provide supplementary material and programs that reproduce
  the results shown in this paper.
\end{abstract}

\section{Introduction}%
\label{sec:Introduction}

Disruption prediction in fusion reactors, engine fault prediction, fraud
detection, and storm surge prediction are just a few exemplary problems from of the
large variety of fields that benefit immensely from anomaly detection in time series.
In this work we will focus on large-scale, high-resolution ocean simulations that
cover the whole Earth with more than 30 different variables such as
temperature, velocity and density easily take up tens of gigabytes for a single
time-step. The vast majority of the simulated ocean, much like the real ocean,
is almost completely unexplored. Unknown physical behaviour hidden in
these data sets could potentially be found by an automated anomaly
detection.  An example of such an anomaly is the bimodal ocean current
called \emph{Kuroshio} on the coast of Japan. In irregular periods of several
years it switches from an elongated to a contracted state. The origin
of this phenomenon is still subject of debate \citep{qiu2000}. A detection of similar
anomalies would be an important finding in itself, but could also contribute to a deeper
understanding of the Kuroshio anomaly and the ocean circulation as a whole.

As we do not wish to restrict the methods to a particular type of anomaly, we must
define what is ``normal'' just by examining the available data. Combined
with the abundance of climate simulation data, this makes neural networks a
promising candidate to solve the problem.
\emph{Echo State Networks} (ESN) have performed
well in predicting low-dimensional chaotic dynamical systems \cite{pathak2017}
and are comparatively easy to train, which enables us to create an \emph{adaptive} anomaly detection framework.
In this work, we extend the method to work well on high-dimensional spatio-temporal data sets, targeting
ocean simulation data.

\subsection{Defining Normality}%
\label{sub:defining_normality}

We aim to create automated anomaly detection algorithms that find contextual anomalies in
large spatio-temporal data sets. We do not assume prior knowledge about the physics that produce the
data, so we do not know in advance the precise nature of anomalies we are looking
for.  This requires that we quantify what is normal, so that whatever
deviates significantly from this can be considered anomalous. Our scheme will
be to identify \emph{normality} with \emph{predictability}: 
if we can build a reliable machinery for predicting future time steps,
the normality of a subsequence can be measured by how well we were able
to predict it in the context of the history preceding it.
Given an input sequence $\matr{U}$ of length $M$
\begin{equation}
  \matr{U} = (\vec{u}_0, \vec{u}_1, ..., \vec{u}_M)
\end{equation}
the prediction problem can be formulated as the search for a model $F$, that
returns a good estimate $\matr{Y}$ of the next $N$ true values $\matr{D} =
(\vec{u}_{M+1}, \vec{u}_{M+2}, ..., \vec{u}_{M+N})$ (further also refered to as \emph{labels}).
\begin{equation}
  \matr{Y}
    = (\vec{y}_{M+1}, \vec{y}_{M+2}, ..., \vec{y}_{M+N})
    = F(\vec{u}_0, \vec{u}_1, ..., \vec{u}_M)
    = F(\matr{U})
\end{equation}
The model $F$ will in our case be based on a type of \emph{Recurrent Neural
Network} (RNN) called \emph{Echo State Network}, which we extended to exploit spatial correlations
in the input as explained in
depth in Sec.~\ref{sub:reservoir_computing}.  The acquired prediction is
subsequently treated as the expected (\emph{normal}) behaviour of the system.
Given a good prediction, detecting anomalies becomes  easy.
The error sequence $\matr{E} = (e_0, ..., e_M)$ can be defined as a distance
between prediction and truth:
$
  e_t = d(\vt{d},\vt{y})
$.
This transforms the \emph{contextual anomaly} detection problem to 
\emph{simple anomaly} detection.
We discuss appropriate error metrics $d(\vt{d},\vt{y})$ in Section \ref{sec:methods}.
The final step is to automatically find good thresholds for the error.
We do this by way of
a \emph{normality score} $\Sigma\in [0;1]$,
which estimates the likelihood that a given
time step is normal:
\begin{equation}
  \label{eq:normality_score}
  \Sigma_t = 1 - \text{erf}\left(\frac{\max\left(0,\mu_m - \mu_n\right)}{\sqrt{2}\sigma_m}\right),  
\end{equation}
Here, $\sigma_m$ and $\mu_m$ are the standard deviation and mean of a sliding window
$(e_{t-m}, ..., e_{t})$ representing \emph{recent history} of the error sequence $\matr{E}$.
The \emph{local mean} $\mu_n$ is calculated
from a shorter window $(e_{t+1}, ..., e_{t+n})$, where $n \ll m$.
If $\mu_n\le \mu_m$,
then $\Sigma_t = 1$ and step $t$ is considered \emph{normal}.
If $\mu_n\gg \mu_m$, i.e., the error is large compared to recent history,
$\Sigma_t \approx 0$, and time step $t$ is likely to be part of a contextual anomaly.
For spatially resolved anomaly detection on image input series,
$\Sigma_t$ can be calculated for localized neighbourhoods.

\subsection{Bimodality of the Kuroshio}%
\label{sub:datasets}

\begin{wrapfigure}{r}{.5\textwidth}
  \vspace{-2\fontsizelength}
  \includegraphics[width=\linewidth]{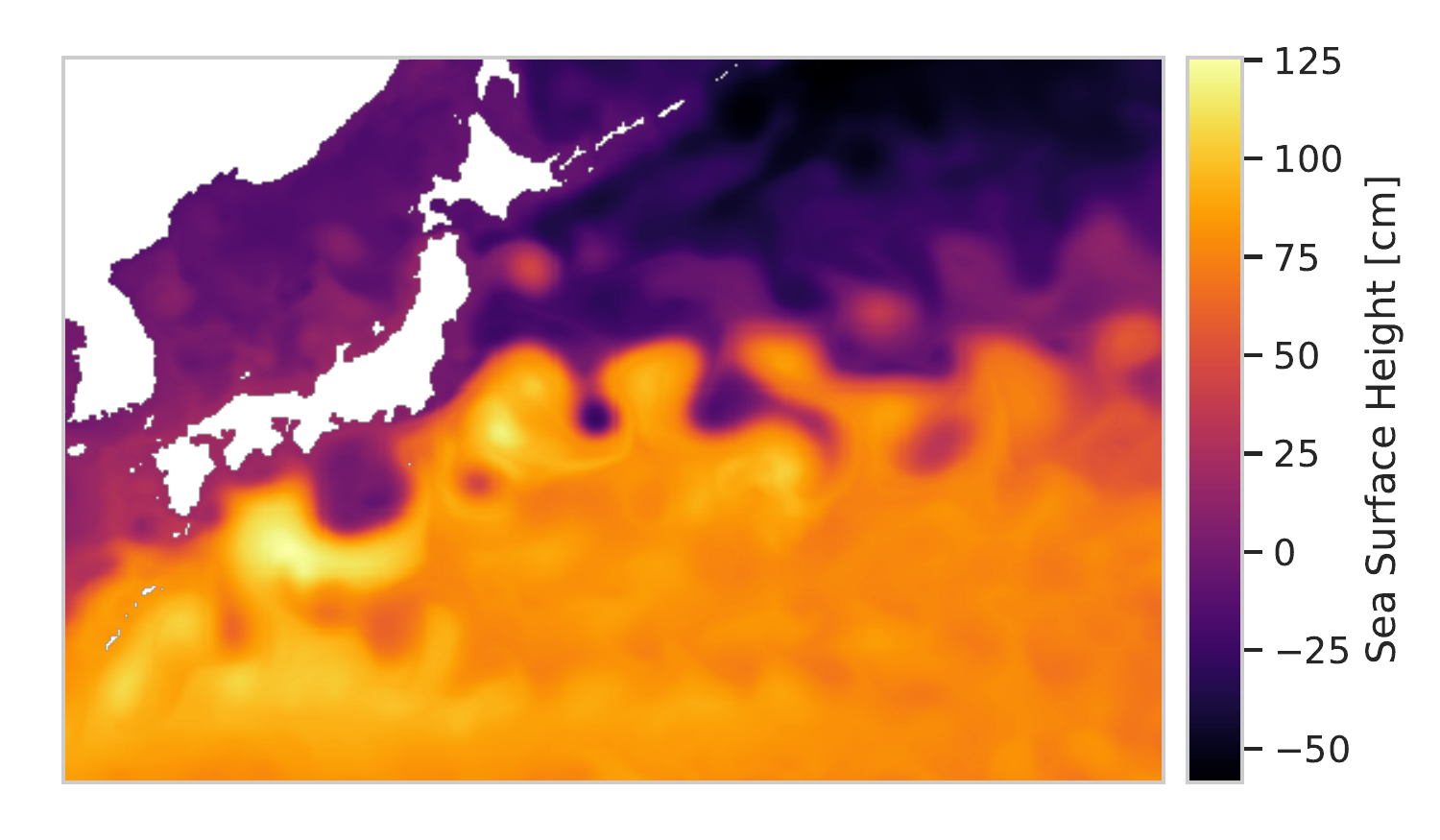}
  \caption{Simulated sea surface heights. The Kuroshio is
    visible as a sharp border flowing along the coast of Japan before
    turning into the North Pacific basin. Shown here is
  a $440 \times 290$ window of the global $3600\times 2400$ domain.}
  \label{fig:intro_kuroshio_snapshot}
  \vspace{-0.7\fontsizelength}
\end{wrapfigure}
\begin{figure}[t]
  \centering
  \includegraphics[width=0.8\linewidth]{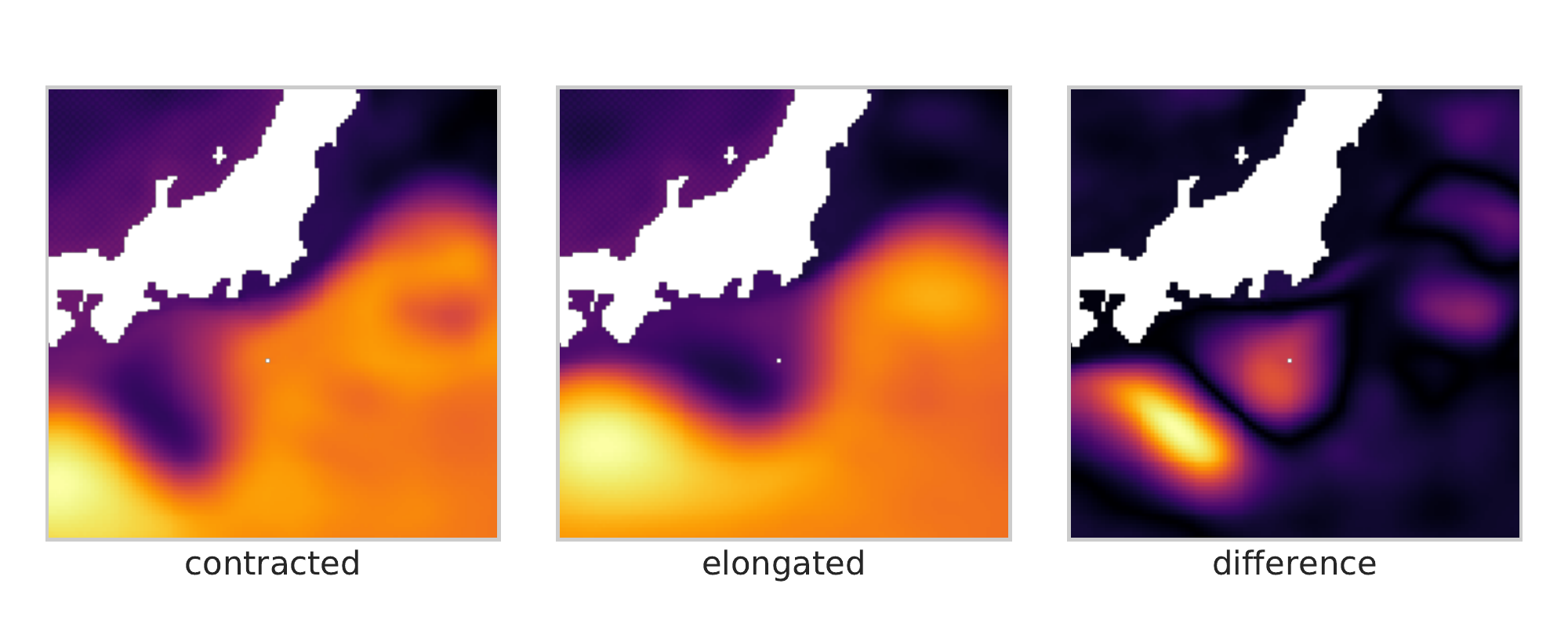}
  \caption{The two distinct states of the Kuroshio created by averaging SSH over
  years on each side of the mode-transition. The images have a size of $100\times 100$ which are sliced out of the global
  simulation domain of $3600 \times 2400$ cells.}
  \label{fig:intro_kuroshio_elon_contr}
\end{figure}

The Kuroshio (Japanese: {\em black tide}) is one of the strongest ocean
boundary currents in the world, and is the result of the western intensification
 \cite{pedlosky2013ocean} of the ocean circulation in the North Pacific. The
3-day mean of simulated Sea Surface Height (SSH) data
(Fig.~\ref{fig:intro_kuroshio_snapshot}) shows the Kuroshio and its extension that
reaches into the North Pacific basin. It carries with it large amounts of
energy, nutrients and biological organisms, which have a strong impact on the
local and global climate and exhibits an interesting and not yet understood bimodality.
In front of the coast of Japan, it oscillates between an elongated and a
contracted state (Fig.~\ref{fig:intro_kuroshio_elon_contr}). The transition
between the two states typically takes one to two years and occurs, as it
seems, randomly every few years. In 2017, it transitioned to its elongated state
for the first time in over a decade, as reported by a Japanese newspaper
\cite{mainichi}.
The simulations that created the SSH data were carried out by \emph{Team Ocean}
at the University of Copenhagen \cite{poulsen2018}.
The \emph{Community Earth System Model} (CESM)
was used to simulate the global domain with a horizontal resolution of
0.1$^\circ$ and 62 depth layers. It writes out 3-day means for all variables,
but in this work only the SSH fields are considered, which results in
images of a total size of $3600\times 2400$ cells. A more detailed
description of the experimental setup can be found in \cite{poulsen2018}.  
As indicated by Fig.~\ref{fig:intro_kuroshio_elon_contr}, the Kuroshio anomaly was
reproduced by the CESM simulations. The two plots of the elongated and the
contracted state are annual means of the
simulation in years exhibiting the two different modalities. Taking the difference of them should give
us an intuition for how a successful anomaly detection should look like.
Detecting the state changes of the Kuroshio with an automated anomaly search
could be the first step in building machinery that can discover novel behaviour in the
vast climate model output data, which is as unexplored as the
oceans of our real world.  Such novelties could, apart from their potential of
displaying new physical processes, contribute to a further understanding
of the behaviour of the Kuroshio itself and the ocean circulation patterns as a
whole. 
Our programs show promise to apply in many other fields, as the methods are quite general.

The algorithms presented in this paper are implemented as an open source
anomaly detection software Python package {\em Torsk}, available on \url{https://github.com/nmheim/torsk}.
All calculations and results can be reproduced by running scripts available
from \cite{supplements}, where we also have made supplementary videos available that
show the dynamical predictions better than is possible on paper.

\section{Methods}%
\label{sec:methods}

We first briefly review \emph{Recurrent Neural Networks}, before
discussing \emph{Reservoir Computing} and \emph{Echo-State Networks},
the restricted type of RNN on which the present work is based. We then
look at how to compute long term and cyclic trends, and separate these
directly computable trends from the more complicated signals.

\subsection{Recurrent Neural Networks}%
\label{sub:recurrent_neural_networks}

Where feed-forward neural networks are state-less and simply pipe
their input through a sequence of layers, {\em Recurrent Neural
  Networks} (RNN) allow cycles in the weights: the network can be any
directed graph, which cannot necessarily be partitioned into layers.
RNN weights describe {\em transition functions} of dynamical systems,
more well suited than FNN for modeling {\em processes} underlying
e.g.~time series.

\paragraph{Notation:}
We partition the RNN nodes into $m$ input nodes $\vec{u}$ and
output nodes $\vec{y}$, and $n$ internal state nodes $\vec{x}$. Their
values at time $t$ are written $\vt{u}$, $\vt{x}$, and $\vt{y}$. The
RNN weights define its time-step transition, which in this work is an
affine transformation of the input and internal state followed by an
activation function:
 \begin{equation}
   \label{eq:recurrent_network_F}
   [\vec{x}_{t+1}, \vec{y}_{t+1}] = \bm{\sigma}\left(\matr{W}[\vt{x},\vt{u}] + \vec{b} \right)
 \end{equation}
 When predicting in free-running mode, the output is fed back as input, $\vt{y} = \vt{u}$.
 The internal state $\vec{x}$ acts as a dynamic short-term memory (STM):
Every new input is mixed in to the previous internal state, gradually
encoding the input sequence into $\vt{x}$.  The length of input sequences
that can be encoded into $\vt{x}$ depends on the STM capacity.  As a rule, the
state size $n$ must be much larger than the input size $m$, in order to create
effective RNN.

The weight matrix $\matr{W}$ is partitioned from $(m+n)\times (m+n)$ into three blocks
$\Win\colon n\times m$, 
$\Whidden\colon n\times n$, and $\Wout\colon m\times (m+n)$, such that
Eq.~\eqref{eq:recurrent_network_F} can be written as
\begin{equation}
  \begin{split}
  \vec{x}_{t+1} &= \bm{\sigma}_h\left(\Whidden\vt{x}+\Win \vt{u} + \vec{b}_h\right), \\
  \vec{y}_{t+1} &= \bm{\sigma}_o\left(\Wout [\vt{u}, \vt{x}]+\vec{b}_o\right)\\
  \vec{u}_{t+1} &= \vt{y}, \text{if predicting}.
\end{split}
\end{equation}
The activation function $\bm{\sigma}$ is written as a vector, allowing a different activation for each
node. While a wide variety of choice is possible, in the present work we will let $\bm{\sigma}_h$ (acting on $\vec{x}$)
be the hyperbolic tangent, and $\bm{\sigma}_o$ (acting on the output $\vec{y}$) be the identity:
\begin{equation}
  \label{eq:state_space}
  \begin{split}
  \vec{x}_{t+1} &= \tanh\left(\Whidden\vt{x}+\Win \vt{u} + \vec{b}_h\right) \\
  \vec{y}_{t+1} &= \Wout [\vt{u}, \vt{x}]+\vec{b}_o
\end{split}
\end{equation}
This simplifies training greatly, as we will see in Section \ref{sub:reservoir_computing}, yet is
sufficiently expressive.

\paragraph{Complications of RNN:}
Training general RNN encounters complications:
The network can be driven through bifurcations in the error surface
during training \cite{doya1993}, which can
prevent the training from converging, as detailed in the appendix.
In addition to convergence-problems, gradient calculation requires
\emph{Backpropagation Through Time} \cite{mozer1995}, similar to full loop-unrolling.
This adds a layer for each time step and quickly leads to excessively deep networks,
prone to vanishing and exploding gradient problems \cite{pascanu2012}
in addition to computational blowup.
\paragraph{LSTM:}
The vanishing gradient problem can be
overcome with network architectures such as the \emph{Long Short-Term Memory} (LSTM) cell
\cite{hochreiter1997}. The LSTM introduces additional forget and input layers,
and has two distinct internal states, the \emph{sigmoid} and \emph{cell} states.
The forget and input layers are trained to decide which parts of a previous sigmoid state are important and
store the information in the cell state. The cell state conserves information 
(and gradient signals) through an arbitrary number time steps by way of a constant self
connection.
Despite the difficulties that arise during LSTM training, they have achieved
remarkable results in a wide range of domains, and
represent the current state of the art in time series forecasting.
However, he most severe problems of RNN training (high complexity
and bifurcations\footnote{Appendix A contains a brief introduction to
bifurcations and demonstrates how they impair RNN training}) unfortunately
still remain for LSTM. 

Our approach goes in a different direction, through a modified version of {\em echo-state networks}.
This avoids the training problems by working with a severely restricted
subset of RNN that can be trained deterministically (and much faster), yet is strong enough for our
purposes. We will use LSTM only to benchmark our methods against.

\subsection{Echo-State Networks and Reservoir Computing}%
\label{sub:reservoir_computing}

The \emph{Echo State Network} (ESN) is a Reservoir Computing (RC) method,
and aims to avoid the problems with
RNN training, while still maintaining the network's temporal awareness.
This is achieved by making a separation between the recurrent part
of the RNN and the subsequent output layer that maps the internal state to the
desired outputs.
Traditionally, the recurrent weight matrices (Eq.\eqref{eq:state_space})
are random projections and are kept constant for all times.  Only the
weights $\Wout$ of the output layer are optimized during the
training phase of the network.
The ESN promises to eliminate all the problems of high computational complexity, vanishing gradients, and
bifurcations during training \cite{jaeger2001}.
It strongly reduces the range of computational processes that can be expressed,
but if this were not the case, efficient training would be out of the question.
A universality result for time-invariant fading-memory filters has been shown by \cite{grigoryeva2018} for similar RC networks,
but the exact computational power of the ESN is not yet established.
In practice we see that Reservoir-RNNs are strong enough to model the very complicated
chaotic and turbulent ocean simulations we study.

Although the reservoir is not optimized at all, it still provides a
non-linear expansion into a higher dimensional space and serves as the
network's short-term memory.
As the goal is to make predictions based on the history of a given
time series, we have to construct an internal state that gradually forgets the
previously seen inputs. An ESN that exhibits this behaviour is said to satisfy the
\emph{echo state property}, traditionally achieved by  
initializing the recurrent weight matrices $\Whidden$ and $\Win$ with a random
uniform distribution $\mathcal{U}(-1, 1)$ and scaling according to two
hyper-parameters: The spectral radius $\rho(\wmatr{})$, and a scaling
factor $\kappa$ for $\Win$.  The spectral radius
determines the influence of the previous internal state on the current one.
The scaling factor $\kappa$ in turn represents the
influence of the current input on the current internal state. 
Finding the right $\rho$ and
$\kappa$ are typical hyper-parameter tuning problems, although
one can make some general considerations to restrict their ranges.  For
example, a $\rho > 1$ increases the non-linearity of the network, but in turn
reduces its STM capacity. STM is maximized at $\rho\approx1$ \cite{farkavs2016}.
This effect is discussed in detail by \cite{jaeger2002},
in practice we found $\rho \approx 1.5$ to work well for
the chaotic dynamical systems we studied.

In this work, we keep the random form of $\Whidden$, but design $\Win$
to make the method better suited for simulation and image data,
described in Section \ref{ssub:spatially_aware_input_map}.

\paragraph{Training:}
Probably the most favorable property of ESN is that they can be
trained using plain linear least-squares optimization in one shot: deterministic
and extremely fast.
With the linear output layer $\Wout$, the predictions of the
network can be written as
\begin{equation}
  \vt{y} = \Wout \bar{\vt{x}},
  \label{eq:esn_lin_out}
\end{equation}
where $\bar{\vt{x}} = [\vt{x},\vt{u}]$ is the internal state
concatenated with the corresponding input.  We can
write the system to solve in terms of the concatenated states
$\matr{X}=(\bar{\vec{x}}_1, ..., \bar{\vec{x}}_T)$
and desired outputs $\matr{D}=(\vec{d}_1, ..., \vec{d}_T)$ as:
\begin{equation}
  \label{eq:esn_lin_out_concat}
   \Wout \matr{X} \simeq \matr{D} 
\end{equation}
To find the optimal weights $\Wout$, we can simply solve the overdetermined
system in Eq.~\eqref{eq:esn_lin_out_concat} via linear least squares:
\begin{equation}
  \Wout = \argmin_{\matr{W}} \norm{\matr{WX}-\matr{D}}_2^2
\end{equation}
equivalent to solving the exact normal equations $\matr{XX}^T\matr{W}^T_{\text{out}} = \matr{XD}^T$.

To avoid over-fitting (which leads to diverging
predictions when feeding the output of the ESN back into the input), it
can be useful to use {\em Tikhonov regularization}, which penalizes large coefficients  \cite{MontgomeryRegression}:
\begin{equation}
  \label{eq:tikhonov}
  \Wout = \argmin_{\matr{W}} \norm{\matr{WX}-\matr{D}}_2^2 + \beta^2 \norm{\matr{W}}_2^2
\end{equation}
This is also a least squares
problem, equivalent to the normal equations
$(\matr{XX}^T+\beta^2\matr{I})\matr{W}^T_{\text{out}} = \matr{XD}^T$.
The least-squares problems are solved directly instead of solving the
normal equations (to avoid squaring up the condition number), and 
the numerical properties have been good for most systems we have studied.
However, for some data series the condition number becomes large,
leading to numerical blowup. For these cases, we have implemented
a slower but highly numerically stable SVD-based least-squares-approximator that
projects on a well-conditioned subspace, ensuring no more
than half the available accuracy is ever lost. 
The choice of optimization method for training is specified by the
user as calculation input.  Effectively, $\beta$ becomes
another hyper-parameter that may need tuning for good
results.

\paragraph{Adaptive Detection:}
The one-shot optimization drastically simplifies not only the training of our
framework, but also the anomaly detection itself, which operates on sliding input
windows. For every window we can find the model that best approximates
the data (in a least-squares sense) within a matter of seconds. This enables us
to retrain the model on the fly resulting in an adaptive outlier detection,
which would be much more complicated to achieve with with e.g.~LSTM.
The amortised computational cost of the optimization step can be reduced one order
by using recursive least squares for the online ESN training.
This is left as future work: our current implementation solves the full least-squares system
in each iteration.

\subsection{Extending the ESN}%
\label{sub:extending_the_esn}

\subsubsection{Spatially Aware Input Map}%

\begin{figure}
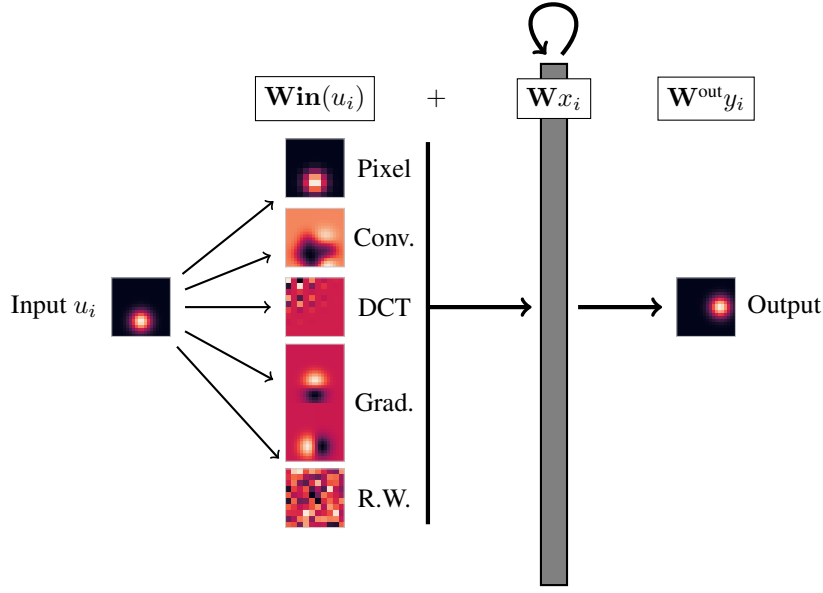

  \centering
  \begin{minipage}{.7\linewidth}
  \ExtendedESN
  \end{minipage}
  \caption{Schematic of the spatial ESN with various input maps.
    The shown input map consists of a downscaled input image, a random convolution,
    a DCT, a spatial gradient and a random input matrix.}
  \label{fig:extended_esn}
  \vspace{-0.5cm}
\end{figure}

Traditional ESN work well for one- and two-dimensional chaotic systems, but when we applied them to
high-dimensional, spatio-temporal data sets, we encountered their limits.
The predicted frames sometimes either quickly
diverge, or converge to what appears to be the mean of the training sequence
and the prediction either stays there, or randomly jumps out of this fixpoint.
We believe this to be caused by the ESN architecture, designed for 1D or few-D systems,
which randomly distributes information from the input frames into the internal state vector,
totally discarding spatial correlation among variables.
We replace the random $\Win$ by a function $\Winf$,
which is a concatenation of smaller input map functions that extract common image features
in the process of mapping to the higher-dimensional hidden space.
These aim to exploit the spatial correlation inherent in simulation data and images.
In our \emph{spatial ESN} (Fig.~\ref{fig:extended_esn}), these maps can be any function from input to hidden state, but
should amplify
information in the input that aid the network in learning.
We implemented five input maps: Resampling the input image
to a certain size, a simple convolution with either random or Gaussian kernels,
a discrete cosine transform (DCT), a spatial gradient of the input image, and the traditional random matrix.
Since all of these are linear maps, $\Winf$ could still be represented by a (large) matrix, and is therefore still an RNN. However, we compute the transforms directly, bypassing the need to store
the $n\times m$ matrix representing $\Winf$, and exchanging the $\bigO{mn}$ matrix-vector
products by a number of linear or $\bigO{k\log k}$-operations (for feature size $k$).
As the dimension of the hidden state must be in the tens of thousands for
the simulation data prediction, this saving is substantial -- in addition to the
method working better.
The flattened, concatenated outputs form the input contribution to the next internal state.
Hence, the spatial ESN state size is not manually defined,
but derived from the output sizes of the chosen processing functions. The
individual input map contributions are scaled to contribute to
the internal state with a similar magnitude. The spatial ESN is illustrated in
Fig.~\ref{fig:extended_esn}.

\label{ssub:spatially_aware_input_map}

\subsubsection{Spatially Aware Loss Function}%
\label{ssub:spatially_aware_loss_function}
The choice of loss function is instrumental in obtaining effective neural networks.
The simplest choice is to use the (possibly weighted) Euclidean distance for each time-step
between the ground-truth $\vec{d}$ and prediction $\vec{y}$,
\begin{equation}
  \label{eq:euclidean-distance}
  \mathcal{L}(\vec{d},\vec{y}) = d(\vec{d},\vec{y})^2 = \norm{\vec{d} - \vec{y}}_2^2
\end{equation}
i.e., the element-wise squared differences, summed over both space and time. This is
a sensible choice if we don't a priori know how the multiple time-series
under analysis relate to each other.
However, for finite-difference simulation data, just as for image data, this will assign
very large errors to images that are nearly identical: Consider, for example, a
high-contrast image shifted a single pixel to one side.
A number of improved metrics have been developed in the image analysis community to solve
exactly this problem, see for example \cite{simard1993, zitova2003, huttenlocher1992}.
We have implemented the IMage Euclidean Distance (IMED) of \cite{wang2005euclidean} in Torsk, which
includes the spatial correlation between pixels/cells by way of a normal distribution over the image coordinate space:
\begin{equation}
  \label{eq:Gij}
  G_{ij} = \frac{1}{2\pi\sigma^2}e^{-\frac{(x_i-x_j)^2+(y_i-y_j)^2}{2\sigma^2}}\\
\end{equation}
where the time steps $\vec{d}_t$ and $\vec{y}_t$ are $M\times N$ images flattened
to $MN$-vectors, and $x_i,y_i$ are the image coordinates corresponding to index $i$.
Then the IMED between two images is
\begin{equation}
  \label{eq:IMED}
  d_{IMED}(\vec{d}_t,\vec{y}_t)^2 = \norm{\vec{d}_t-\vec{y}_t}_G^2
  = (\vec{d}_t-\vec{y}_t)^T \matr{G} (\vec{d}_t-\vec{y}_t)
\end{equation}
that is, $\matr{G}$ is an $MN\times MN$ linear transformation that mixes
pixel/cell-values within their spatial vicinity. 
The IMED is the Euclidean distance between $\matr{G}^{1/2}$-transformed images:
\begin{equation}
  \norm{\vec{z}}_G^2 = \vec{z}^T \matr{G} \vec{z} = \norm{\matr{G}^{1/2}\vec{z}}_2^2
\end{equation}
so that the IMED loss function
\begin{equation}
  \label{eq:L_IMED}
  \mathcal{L}_{IMED}(\vec{d},\vec{y}) = \sum_{t=t_0}^{t_n} \norm{\vec{d}_t-\vec{y}_t}_G^2
  = \sum_{t=t_0}^{t_n} \norm{\matr{G}^{1/2}(\vec{d}_t-\vec{y}_t)}_2^2 
\end{equation}
is minimized simply by solving a $\matr{G}^{1/2}$-transformed linear least-squares
system, and  can be subjected to
Tikhonov regularization to ensure small coefficients and prevent over-fitting
in the same way as the ``flat'' Euclidean distance.
Hence, the IMED is incorporated into ESN-learning simply by transforming
the labels by $\matr{G}^{1/2}$ and optimizing as usual.
Since $\vec{y}_t = \matr{G}^{1/2}\matr{W}^{out}\vec{x}_t$, we find
\begin{equation}
  \begin{split}
    \mathcal{L}_{IMED}(\vec{d},\vec{y}) &= \sum_{t=t_0}^{t_n} \norm{\matr{G}^{1/2}\vec{d}-\matr{G}^{1/2}\matr{W}^{out} \vec{x}_t}_2^2\\
    &= \sum_{t=t_0}^{t_n} \norm{\matr{G}^{1/2}\vec{d}_t-\matr{W}^{out}_G \vec{x}_t}_2^2
  \end{split}
\end{equation}
whereby a linear least-squares or Tikhonov minimization with fixed $\vec{d}$ and $\vec{x}$
yields the optimal $\matr{W}^{out}_G \equiv \matr{G}^{1/2} \matr{W}^{out}$.
To recover the output
weights that predict the expected, non-transformed images,
we transform back as $\matr{W}^{out} = \matr{G}^{-1/2}\matr{W}^{out}_G$.

Computing $\matr{G}^{1/2}$ requires a diagonalization,
which at first sight is $\bigO{(MN)^3}$ and prohibitively expensive,
but it can be efficiently implemented by seperation of variables:
\begin{equation}
  \begin{split}
    \matr{G}_{ij} &= \frac{1}{2\pi\sigma^2}e^{-\frac{(x_i-x_j)^2+(y_i-y_j)^2}{2\sigma^2}}\\
  &= \frac{1}{\sqrt{2\pi\sigma^2}} e^{-\frac{(x_i-x_j)^2}{2\sigma^2}}
  \frac{1}{\sqrt{2\pi\sigma^2}} e^{-\frac{(y_i-y_j)^2}{2\sigma^2}}\\
  &= g\!\left(x_i-x_j \mid \sigma^2 \right)\,
     g\!\left(y_i-y_j \mid \sigma^2 \right)
  \end{split}
\end{equation}
That is, the $MN\times MN$-matrix $\matr{G}$ is
a Kronecker product $\matr{G} = \matr{G}^x \otimes \matr{G}^y$ with $\matr{G}^x\colon M\times M$
and $\matr{G}^y\colon N\times N$.
The corresponding eigenvalues are $\lambda_{ij} = \lambda^x_i \lambda^y_j$
with $\lambda^x_i$ the $M$ eigenvalues of $\matr{G}^x$ and $\lambda^y_j$ the $N$ eigenvalues
of $\matr{G}^y$. Similarly, the eigenvectors are $\vec{e}_{ij} = \vec{e}^x_i\otimes \vec{e}^y_j$.
Hence, calculating $\matr{G}^{1/2}$ is a $\bigO{M^3+N^3}$ operation (needed only once for
any fixed image size), and application of the transformation is $\bigO{M^2+N^2}$,
as each axis can be transformed independently of the other.

For time-series (time simulation data or video), we will often want to include
the temporal correlation. The separation makes it straight-forward to do this, by transforming
the whole space-time volume
\begin{equation}
  \matr{G}_{ij} =
    g\!\left(x_i-x_j \mid \sigma_x^2 \right)\, g\!\left(y_i-y_j \mid \sigma_y^2 \right)
    g\!\left(t_i-t_j \mid \sigma_t^2 \right)
\end{equation}
with $i,j$ ranging over $1,\ldots,MNT$ for $T$ time-steps.
However, for long time-series, some extra steps are needed to make it efficient.
The present work includes only the spatial IMED; time correlation is delegated to future work.

\subsection{Long term trends and cyclic behaviour}
\label{sec:cycles}

\begin{figure}
  \centering
  \begin{tabular}{ccc}
    \begin{minipage}{.3\linewidth}
      \includegraphics[width=5cm]{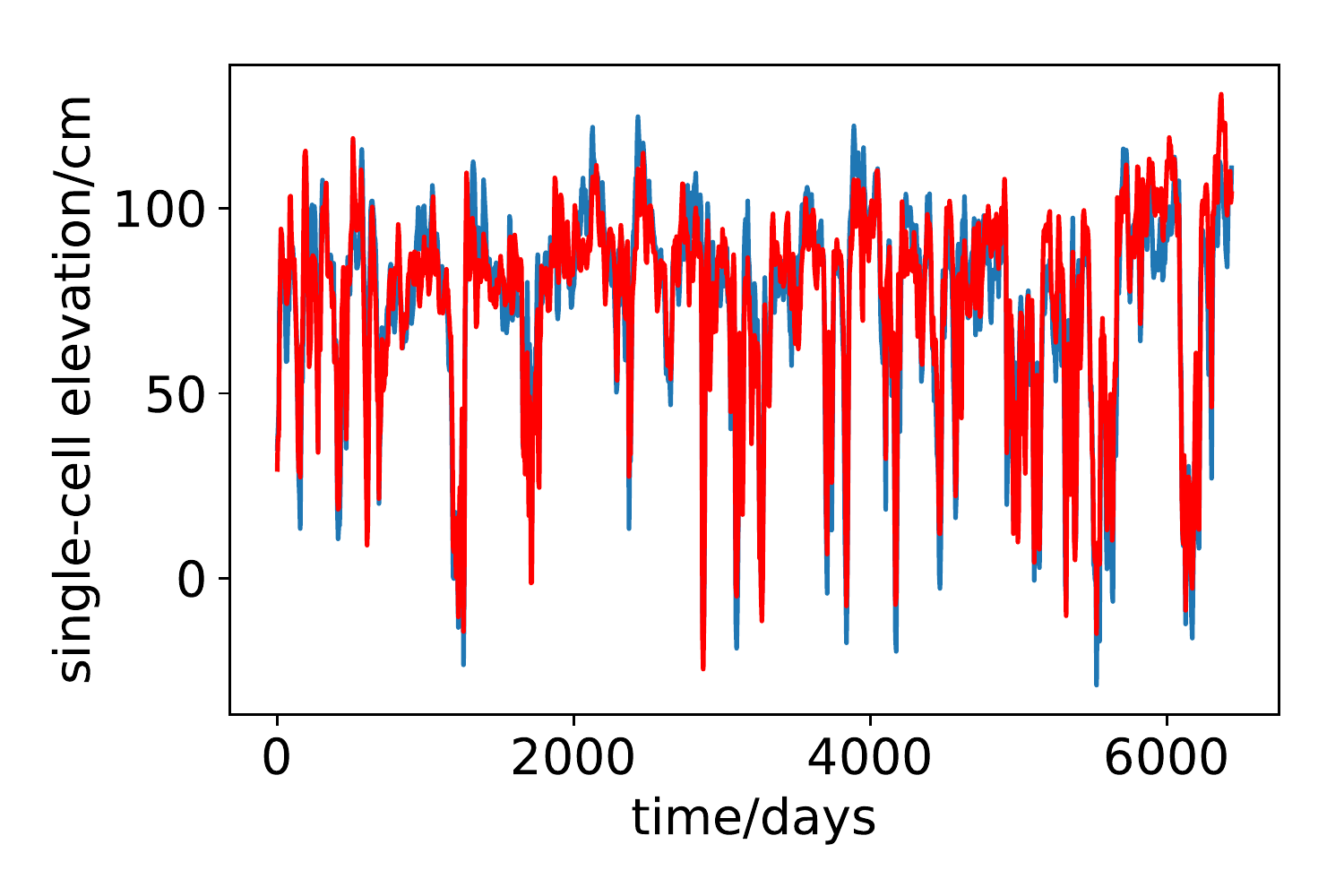}
    \end{minipage}
    &
    \begin{minipage}{.3\linewidth}      
      \includegraphics[width=5cm]{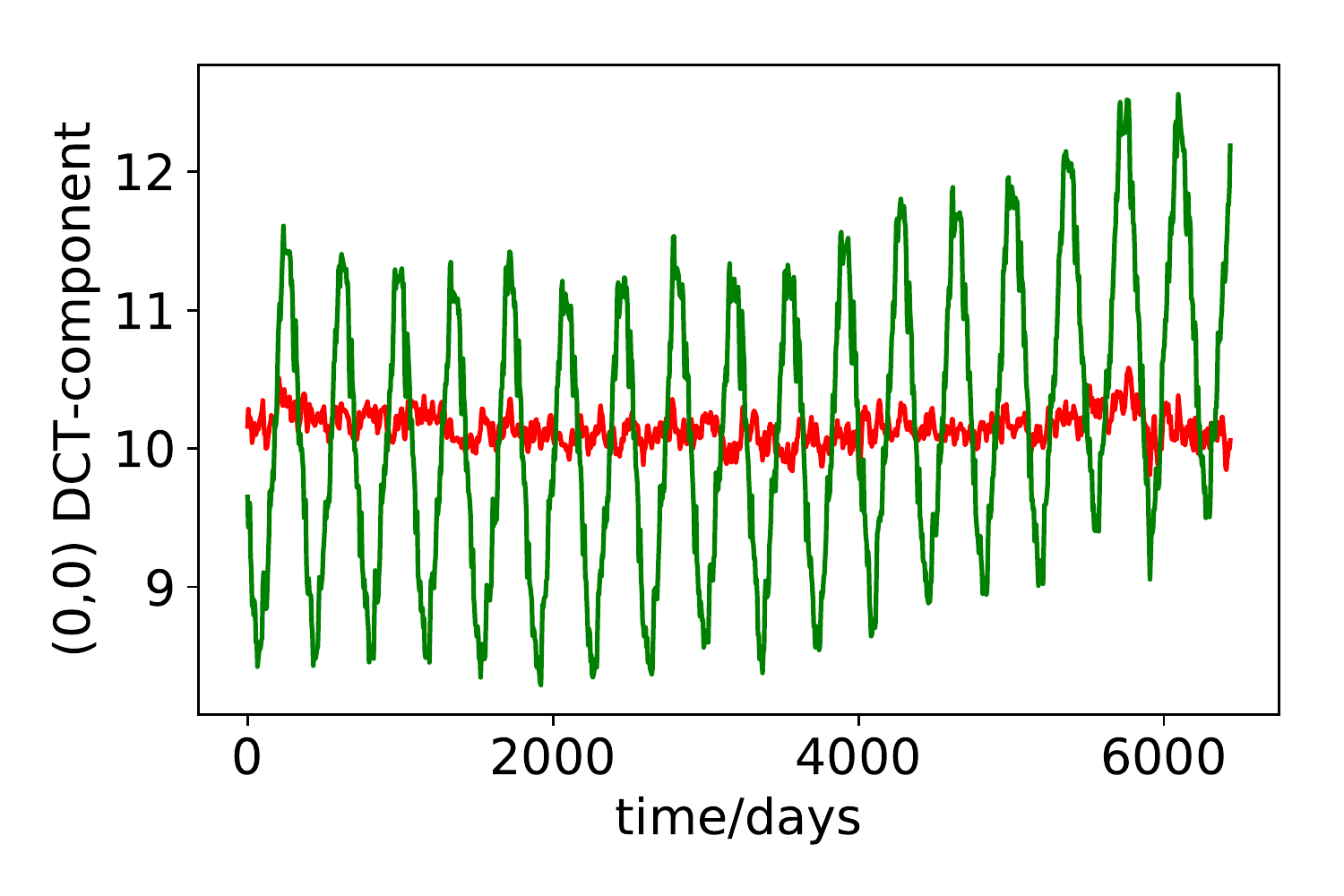}
    \end{minipage}
    &
    \begin{minipage}{.3\linewidth}      
      \includegraphics[width=5cm]{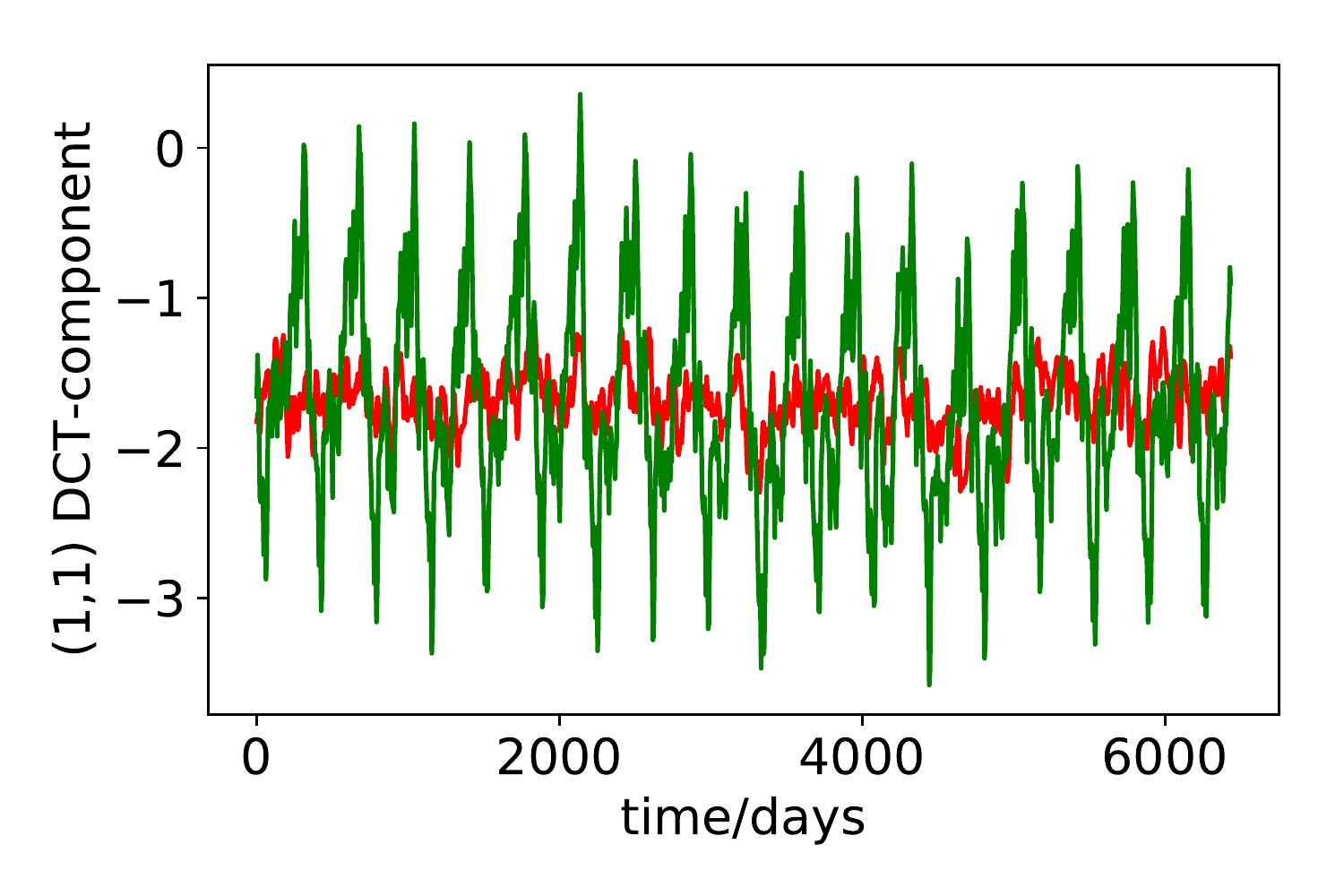}
    \end{minipage}
     \\
    (A)&(B)&(C)\\        
  \end{tabular}
  \caption{Time series shown together with their
    de-trended counterparts
    for (A) single pixel, (B) the (0,0)
    DCT component, and (C) the (1,1) DCT component.
    Higher frequencies are less-and-less well described
    by yearly cycles, as can be seen in \cite{supplements}.
  }
  \label{fig:cycle-pixel}
\end{figure}

Many time series are driven by cyclical forcings, resulting in simple long term trends and ``seasonal'' variations that can
be analysed separately from a much smaller chaotic or turbulent component.
In the case studied here, ocean behaviour is strongly driven by the annual cycle of Earth orbiting the Sun.
The seasonal behaviour resulting from this can be analysed directly without the need for neural networks, as will be described below, leaving
a much stronger signal of the difficult chaotic component.
However, for individual simulation cells (or image pixels),
the seasonal component is obscured by local turbulence,
as seen in Figure \ref{fig:cycle-pixel}(a). Instead, it is large-scale features that are seasonal.
The lowest frequency components in a spatial cosine transformation are extremely well-described
by an average yearly cycle on top of a long-term quadratic trend. Examples are shown 
in Figure \ref{fig:cycle-pixel}(b) and (c).

Separating out the trend and cyclic components lets our neural network
machinery focus on the signals that are important for anomaly
prediction.  However, it also provides a baseline method for
prediction: given a starting point, we can simply continue along the
average cycle added to the long term polynomial trend.  We will use
this as a benchmark against which to assess our neural networks' predictive
accuracy.

We describe how to do this for a 1D time-series $\vec{f} = [f(t_1),\ldots,f(t_n)]$ given a known cycle length $l_C$: a full image is processed
by applying the same procedure independently to each variable to be detrended.
In our case, we DCT-transform the spatial domain (i.e., 2D-DCT every time-step) and detrend each component.
\paragraph{Decomposition:}
First, a $d$-degree polynomial trend $p(t) = b_0 +b_1 t + \cdots b_d t^d \simeq f(t)$ is computed by a least-squares fit of the entire training data (with $d$ small: 1, 2, or 3):
\begin{equation}
  \left[\begin{smallmatrix}
      t_1^0 & \cdots & t_1^d\\
      \vdots & \ddots & \vdots\\
      t_n^0 & \cdots & t_n^d
    \end{smallmatrix}\right]
  \left[
    \begin{smallmatrix}
      b_0\\\vdots\\b_d
    \end{smallmatrix}
  \right]
  \simeq
  \left[
    \begin{smallmatrix}
      f_0\\\vdots\\f_n
    \end{smallmatrix}
  \right]
\end{equation}
This long-term overall trend $p(t)$ is subtracted from $\vec{f}$
before computing the average seasonal cycle.  In general, the cycle
length $l_C$ may not be an integer; for example, a year is 365.24
days.\footnote{The CESM simulation data works with exact 365-day
  years, but 3-day time-steps. We solve this by rescaling to 5-day
  time steps, i.e., $a_C = 3/5$. } In this case, we first scale the
time by a factor $a_C$ so that the cycle length $L_C = a_C l_C$ in the
new time scale is an integer, and resample $\vec{f}$ smoothly onto the
new $N$ time steps using an $n$-point forward DCT followed by a
zero-padded $N = a_C n$ point inverse DCT.  The time series comprises
$N_C = \floor{\frac{N}{L_C}}$ full cycles, and the average cycle is
found simply by reshaping it into an $N_C \times L_C$ matrix and
averaging over the rows (disregarding the final
$N-\floor{\frac{N}{L_C}}L_C$ elements not part of a full cycle).  The
time series can then be represented as a tuple
$(\vec{\tilde{f}},\vec{b},\vec{C})$, where $\vec{\tilde{f}}$ is the
de-trended time series (of length $n$), $\vec{b}$ the $d+1$ polynomial coefficients of the
long-term trend, and $\vec{C}$ the mean cycle (of length $L_C$).

\paragraph{Reconstruction:}
Given a trend-decomposed tuple $(\vec{\tilde{f}},\vec{b},\vec{C})$,
the original time series $\vec{f}$ is recovered by 1) resampling $\vec{\tilde{f}}$
to the $N$-timescale as described above, 2) adding $p(t)+C_{(t\bmod L_C)}$, and 3) resampling back to the $n$-timescale.
Of course, if the cycle length is already an integer in the original time series, only Step 2 is needed.

\begin{figure}
  \centering
  \begin{tabular}{cccc}
    \begin{minipage}{.22\linewidth}
      \includegraphics[width=1.15\linewidth]{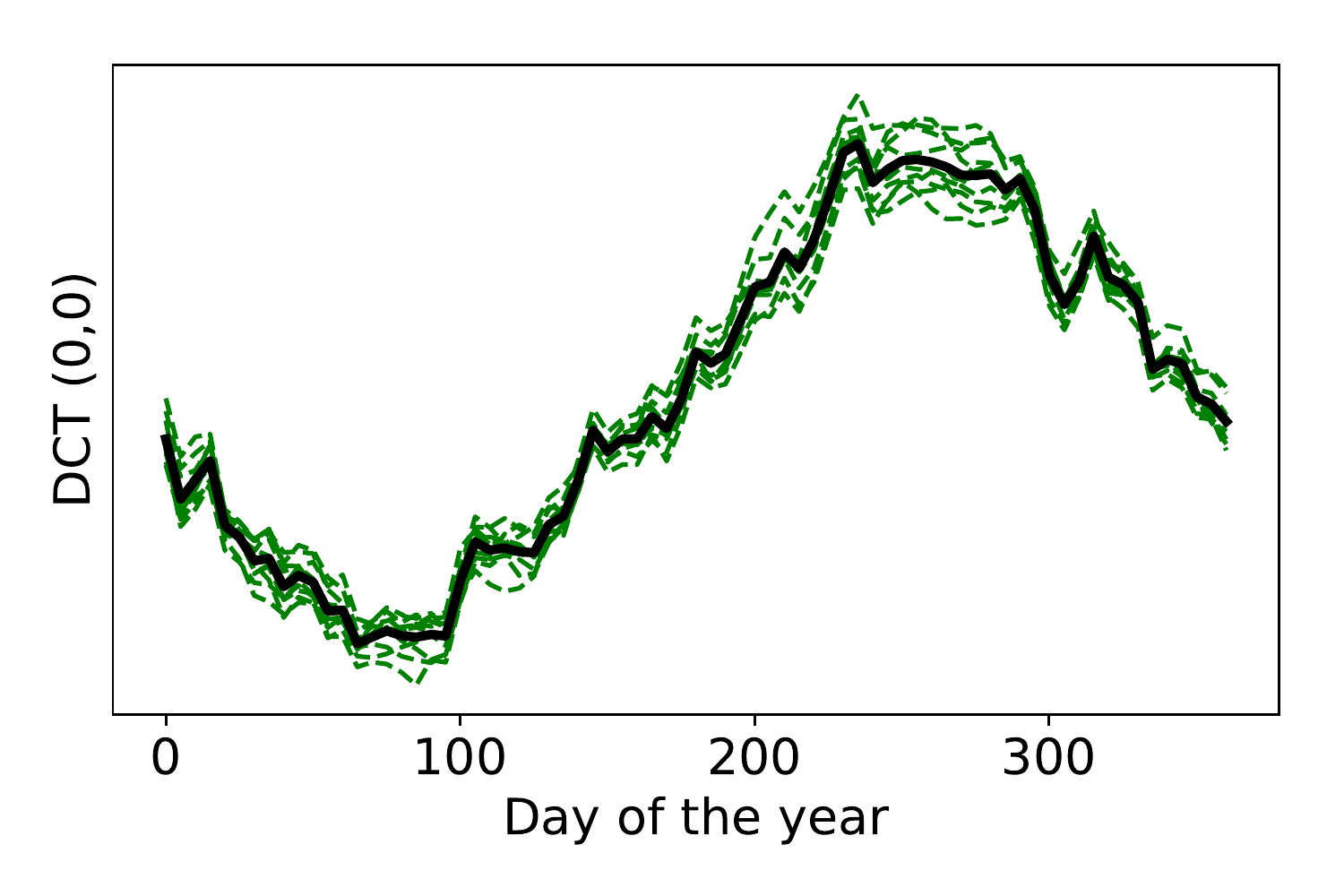}
    \end{minipage}
    &
    \begin{minipage}{.22\linewidth}
      \includegraphics[width=1.15\linewidth]{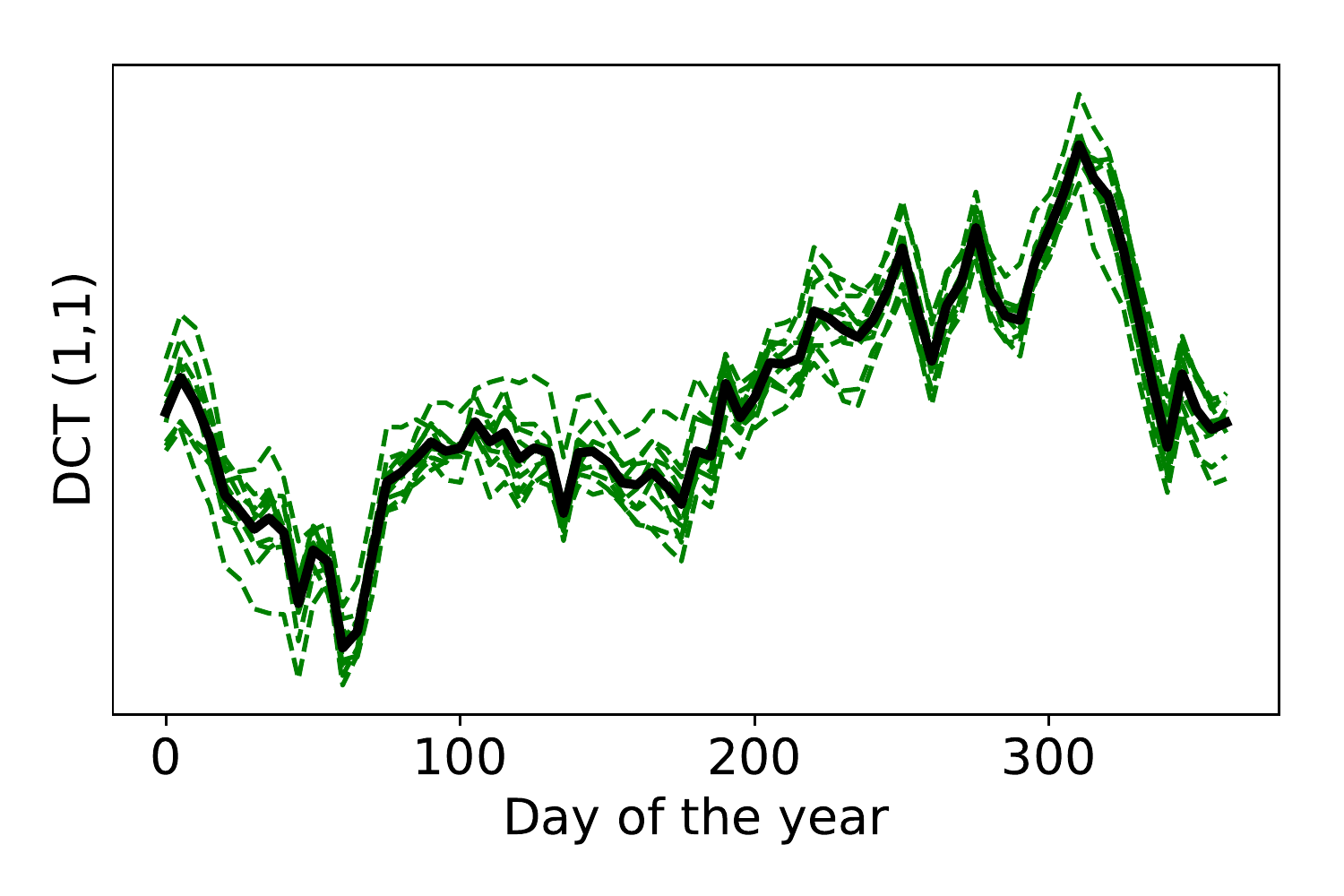}      
    \end{minipage}      
    &
    \begin{minipage}{.22\linewidth}
      \includegraphics[width=1.15\linewidth]{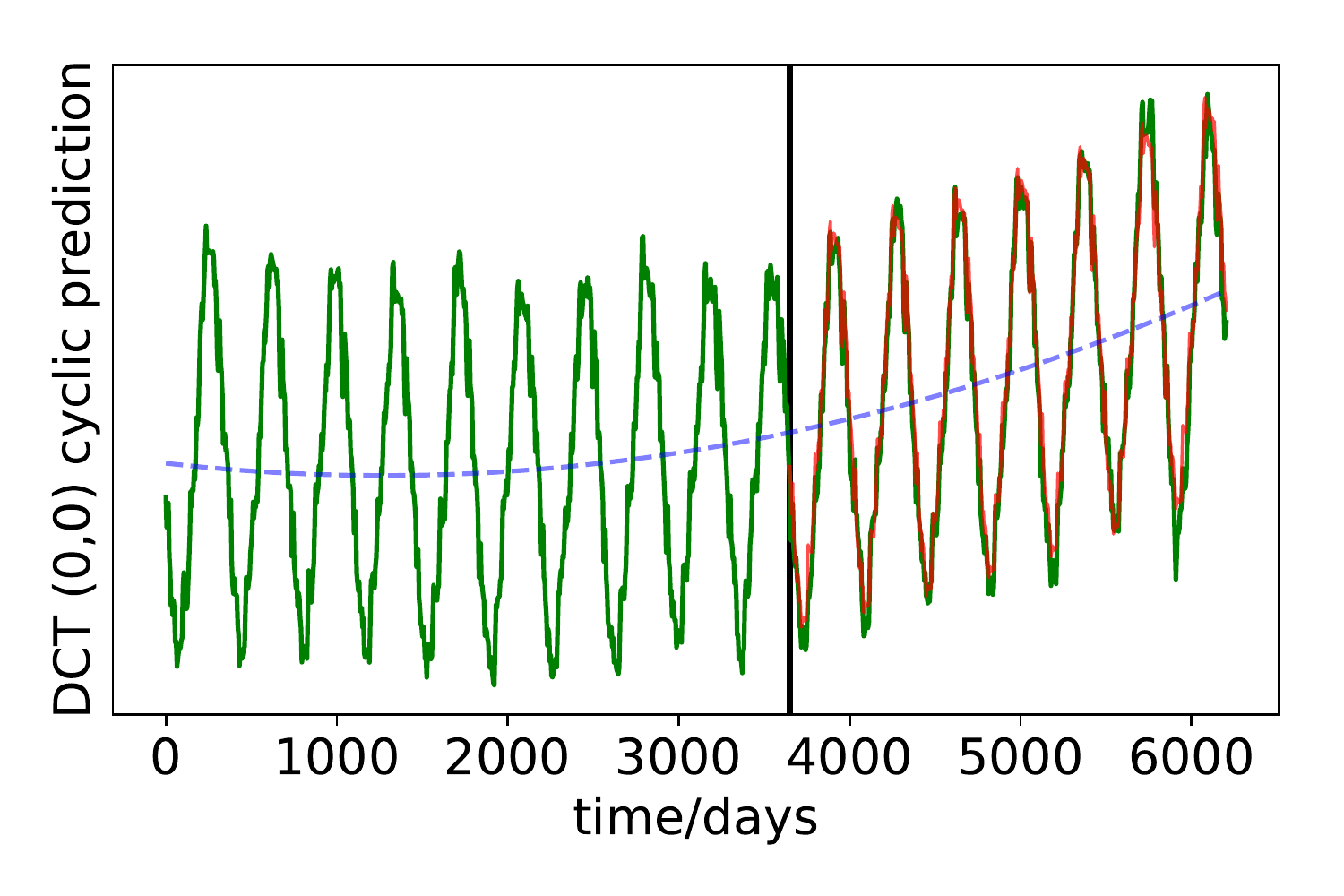}            
    \end{minipage}      
    &
    \begin{minipage}{.22\linewidth}
      \includegraphics[width=1.15\linewidth]{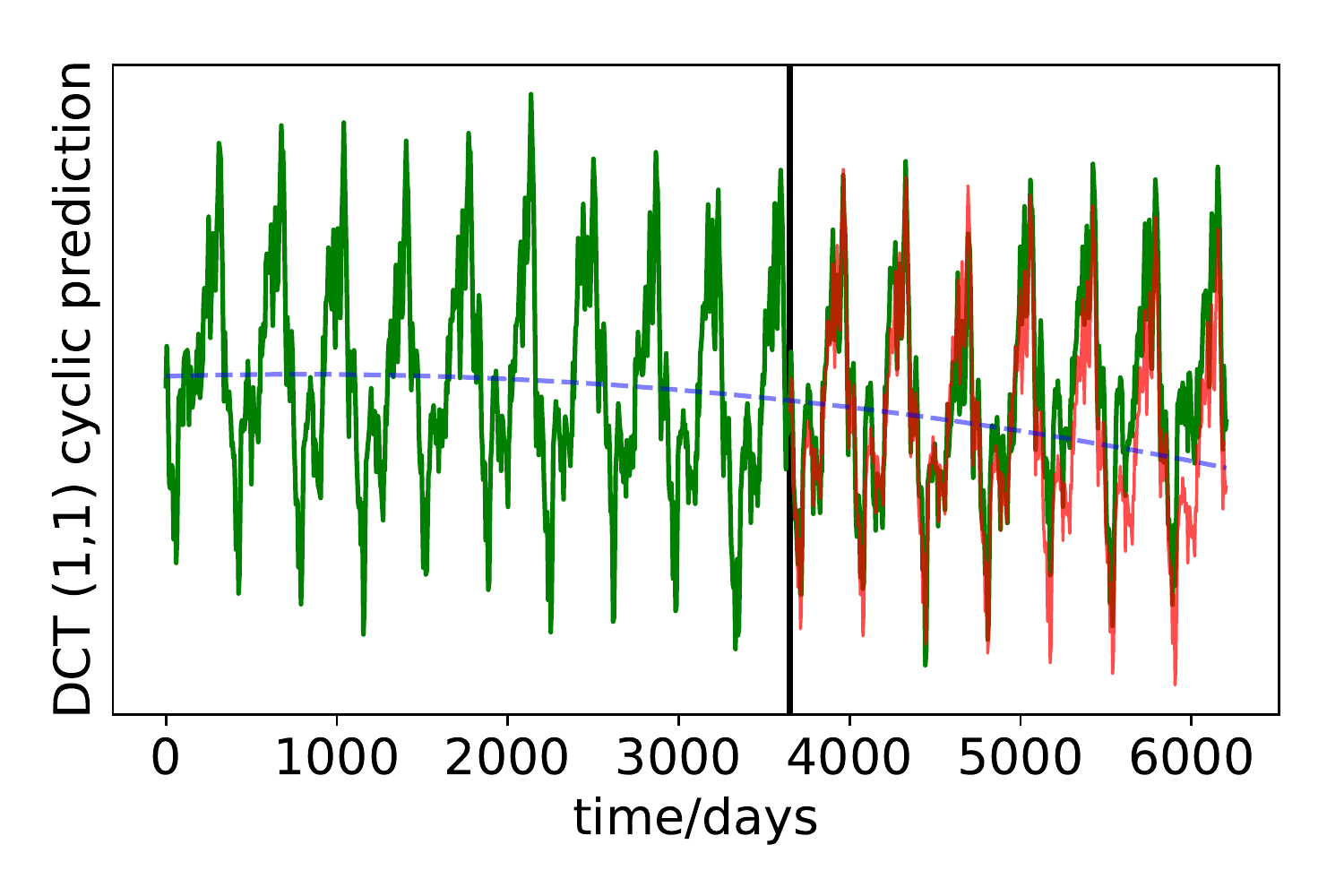}                  
    \end{minipage}
    \\
    \multicolumn{2}{c}{(A)}
    &
    \multicolumn{2}{c}{(B)}      
  \end{tabular}
  \caption{Trends computed from simulation Year 1-10, used to predict Year 11-17.
    (A) Mean cycles and actual cycles for Years 1-10.
    (B) Actual data (green) and predicted (red).
  }
  \label{fig:cycle-predict}
\end{figure}

\paragraph{Prediction:}
Using the trend and mean cycle, we already obtain a quite decent method
for predicting future behaviour simply by continuing the trend
from a starting time $t_0$: 
\begin{equation}
  \label{eq:cyclic-prediction}
  \begin{split}
  \tilde{f}_{t_0} &= f(t_0) - C_{(t_0 \bmod L_C)} - p(t_0)\\
  f(t)           & \simeq \tilde{f}_{t_0} + p(t) + C_{(t \bmod L_C)}
  \end{split}
\end{equation}
corresponding to reconstruction with constant $\tilde{f}_t = \tilde{f}_{t_0}$ in the prediction range.
Fig.~\ref{fig:cycle-predict} shows this method applied to the $(0,0)$ and $(1,1)$
DCT-components of the Kuroshio ocean surface height data. Note that
higher frequencies become increasingly dominated by turbulence;
the supplementary material contains the full calculation.
We use this method as a benchmark against which to evaluate
our neural network prediction methods in Section \ref{sec:results}.

\section{Results}%
\label{sec:results}
In this section, we benchmark our anomaly detection framework.
Starting by ensuring that anomaly detection framework works for the
one-dimensional chaotic Mackey-Glass system (MG), we gradually increase the
difficulty of the prediction task towards the high-dimensional ocean simulation
data set.  We show that we can outperform trivial, cycle-based, and even LSTM predictors in
chaotic systems without prior knowledge of the underlying physics of the data.

The approach is the same throughout: In each prediction iteration, a
number $L_{trans}+L_{train}$ of input frames are fed to the network to generate
internal states. The first few $L_{trans}$ states are discarded to get rid of
transient effects of the initial state, and the remaining $L_{train}$
states are used for training the output layer as described in
Sec.~\ref{sub:reservoir_computing}.  Now the network can predict the next
$L_{pred}$ steps by feeding the output back into the input of the network. This
process is repeated until the sliding window of $L_{trans}+L_{train}+L_{pred}$
frames has passed over the whole data set.
We will refer to this approach as {\em online} ESN, because the output
layer is re-optimized continuously as the sliding window moves over the data, such that it predicts
the frames that come directly after the training sequence.
Transient length and spectral radius $\rho$ are of course tightly
coupled, as a smaller $\rho$ results in shorter memory retention and makes a smaller
$L_{trans}$ possible. We set $\rho=1.5$ to make the reservoir sufficiently non-linear
and found $L_{trans}=200$ to be sufficiently long to eliminate transient effects.
The sparsity of the reservoir matrix was set to 90\%.
$L_{pred}$ should be set with respect to the prediction performance on the individual
data set. 
In addition, the prediction length has to be chosen long enough such that the
error sequence becomes sufficiently anomalous when unexpected behaviour is encountered,
but also short enough that short anomalies are not averaged out by correct
predictions. The detection will therefore work best for anomalies with length a few times $L_{pred}$.

For each iteration, we compute the prediction error and perform the final
anomaly detection on the resulting error sequence $\matr{E}$ by calculating the
normality score $\Sigma_t$.  For the normality score we need to set a large
window size $m$ and a small window size $n$ (as described in
Sec.~\ref{sub:defining_normality}).  Throughout this paper we use $m=100$ and
$n=5$ unless stated otherwise.
Some reasonable defaults for the input maps are shown in Fig.~\ref{tab:input_maps}.

We benchmark the performance of the ESN predictions against the
cycle-based prediction described in Section \ref{sec:cycles},
as well as to an LSTM. In addition, we 
compare it as a sanity-check to the trivial
prediction, which just constantly predicts the last value of the training sequence.
The LSTM is trained only on the first training data of length  $L_{train}$, as online training
with LSTM would be too resource-intensive. For each prediction, the trained LSTM was then fed input frames until
prediction start, then switched to feeding output back as input for the $L_{pred}$ prediction steps.
While we use online ESN for actual anomaly detection, we also compare the to {\em offline} ESN, also
only trained once on the initial training set, to make the comparison with the LSTM more clear.

\subsection{Mackey Glass}%
\label{sub:mackey_glass}
\begin{figure}
  \centering
  \includegraphics[width=0.57\linewidth]{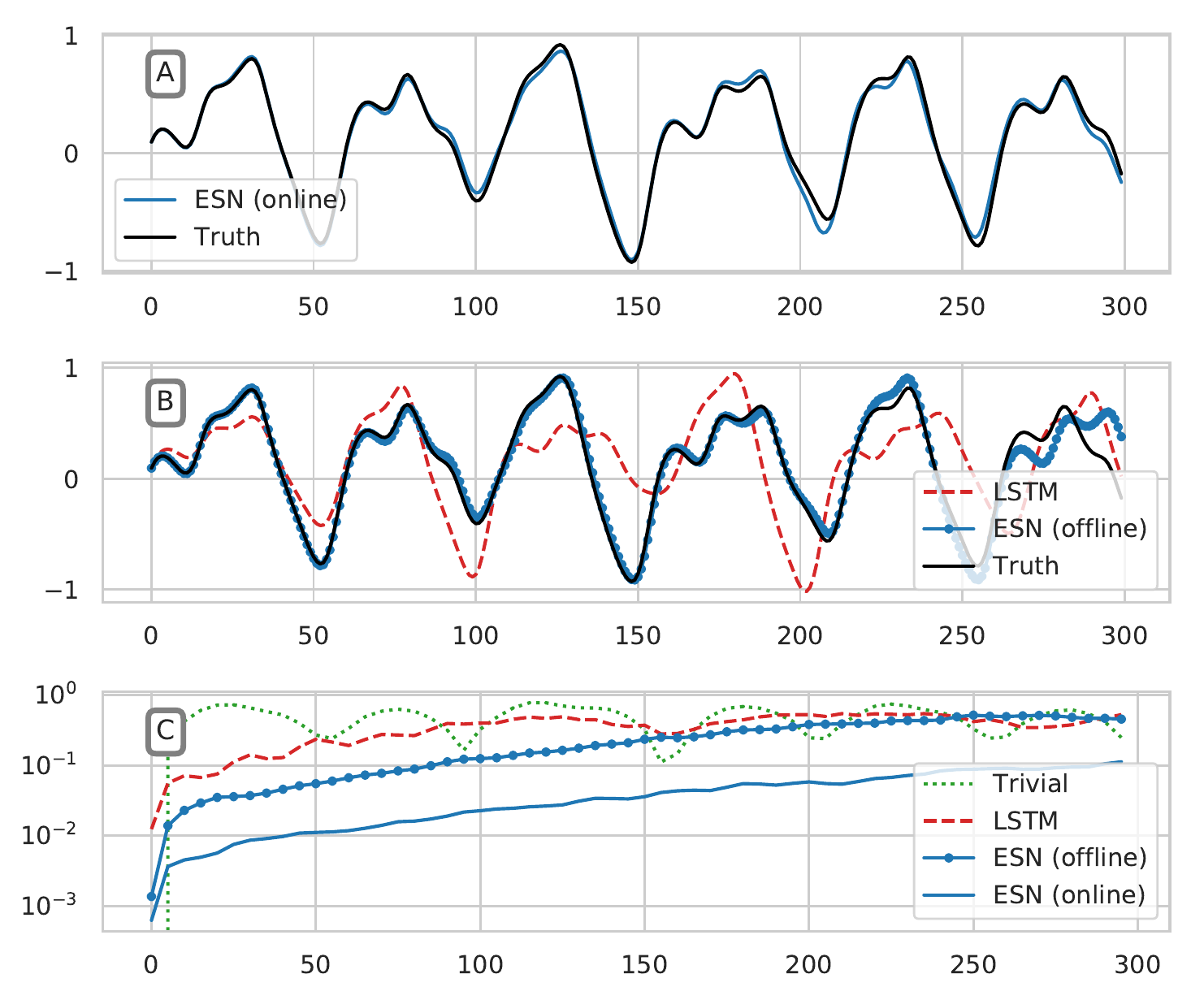}
  \caption{Comparison of ESN and LSTM. (A) Predictions on the MG system with online ESN (trained just before prediction start).
    (B) Predictions with compared methods (trained 500 steps before prediction start).
    (C) Mean prediction error over 100 sliding window iterations.
    Both online and offline ESN outperformed
    the LSTM in both prediction accuracy and computation time. It takes about 300 seconds
    to train the LSTM vs. less than one second to train the ESN.}
  \label{fig:res_esn_lstm_1dmackey}
\end{figure}

\begin{figure}
  \centering
  \includegraphics[width=.57\linewidth]{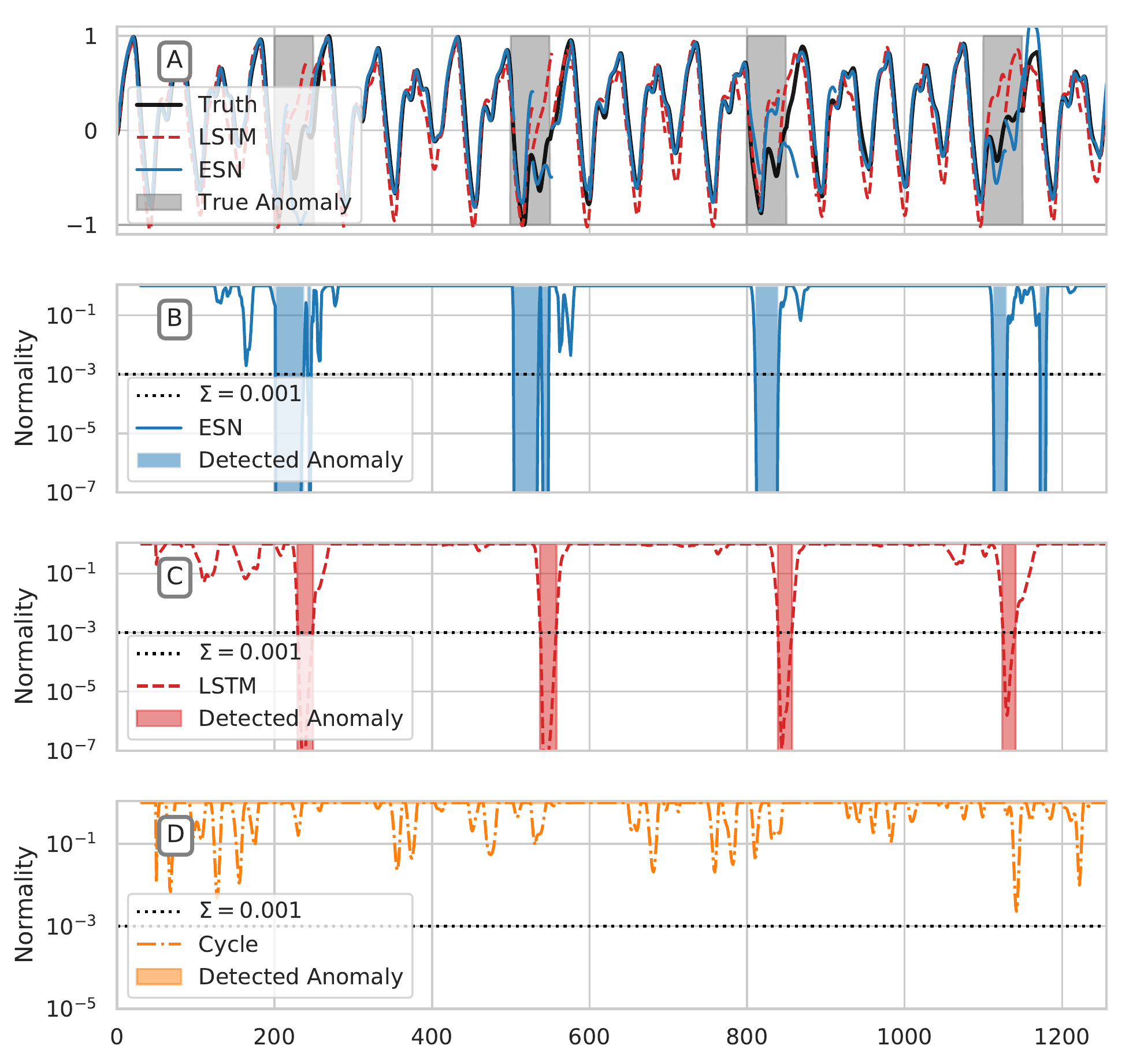}
  \caption{
    (A) 1D MG system with anomalies and exemplary online ESN and offline LSTM
        predictions 25 steps ahead.
    (B) Online ESN normality score and detected anomalies shaded in blue.
    (C) Offline LSTM normality score and
    (D) Cycle-based normality score.
  Points are classified as anomalies when $\Sigma < 0.001$.
  }
  \label{fig:res_1dmackey}
\end{figure}

The Mackey-Glass (MG) system is a simple delay-differential equation
that exhibits chaotic behaviour under certain conditions, defined as
\begin{equation}
  \label{eq:mackey_glass}
  \frac{\partial x}{\partial t} = \beta \frac{x_\tau}{1+x_\tau^n} - \gamma x,
\end{equation}
where $\beta$ and $\gamma$ are constants and $x_\tau$ denotes the value
$x(t-\tau)$, representing the delay.
The system is studied
extensively in non-linear dynamics and serves as a benchmark for chaotic
prediction algorithms. 
We will use the MG system to construct
example tasks that builds up our methods towards finally
applying it to the ocean simulation data.

Fig.~\ref{fig:res_esn_lstm_1dmackey}A shows the true time series
together with a single 300-step prediction using ESN and LSTM, respectively, both
with hidden state size 1000.
The networks were trained on a training sequence of length $L_{train}=2000$.
For the LSTM this sequence is subsampled randomly into batches of size 32 with
a subsequence length of 200.
It took approximately {\bf 300 seconds} to train the LSTM, while the ESN is optimized
in less than {\bf one second}. Plot~\ref{fig:res_esn_lstm_1dmackey}B shows that
despite the reduced complexity of the ESN, its prediction error is lower than
the LSTM's. This is likely because the ESN always finds the exact optimum for its output layer
directly though linear least squares, while the LSTM is optimized via gradient descent and may
get stuck in a local minimum, and in
addition has to overcome the inherent RNN difficulties that were described in
Sec.~\ref{sub:recurrent_neural_networks}.
Finally, the low computational complexity of the ESN makes it possible to train
it online on a moving window and then predict the frames that come immediately
after the training sequence. This results in even better prediction performance
and makes an online, adaptive anomaly detection possible.

To simulate anomalies, we slightly change one
parameter of the MG equation from $\gamma = 0.10$ to $\gamma = 0.13$ for
50 steps during the integration. Time periods where $\gamma = 0.13$ are
shaded in gray in Fig.~\ref{fig:res_1dmackey}A. The resulting normality
sequences for ESN, LSTM, and cycle-based predictions are
shown in Fig.~\ref{fig:res_1dmackey}B, C, and D.
For detection, we set the prediction length to $L_{pred}=25$ (half the anomaly length).
We classify a point as anomalous if $\Sigma < 0.001$.
The ESN reliably detects all anomalies at the correct times (shaded regions),
the LSTM finds them (but a little late), and
the cycle-based prediction is not close. 

\subsection{Lissajous Figures}%

\begin{figure}
  \centering
  \begin{minipage}{.36\linewidth}
    \includegraphics[width=\linewidth]{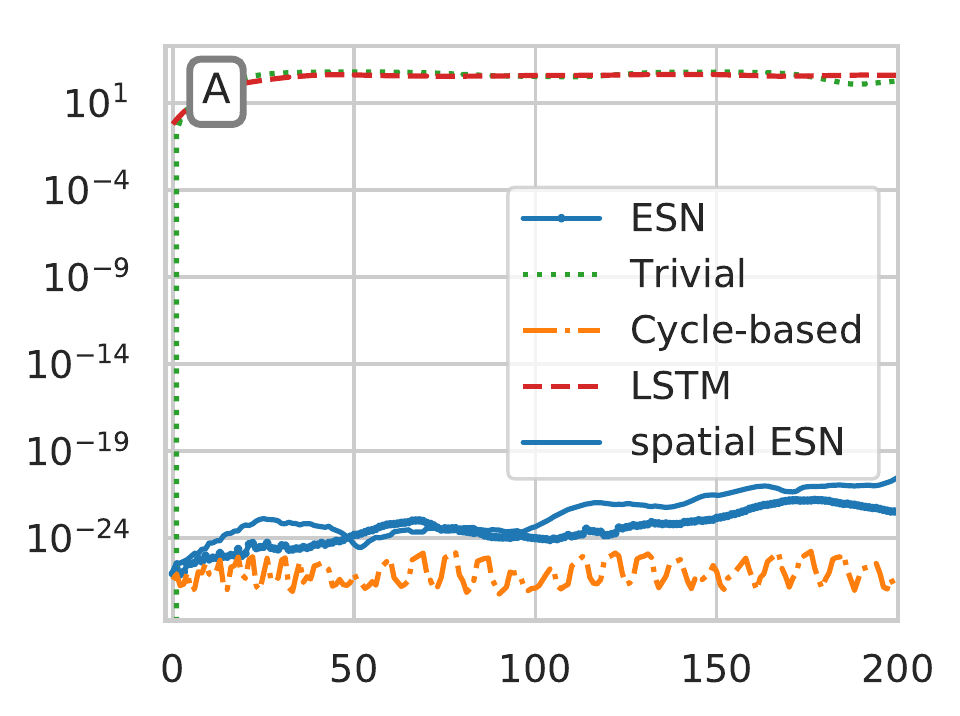}
  \end{minipage}
  \begin{minipage}{.265\linewidth}
    \includegraphics[width=\linewidth]{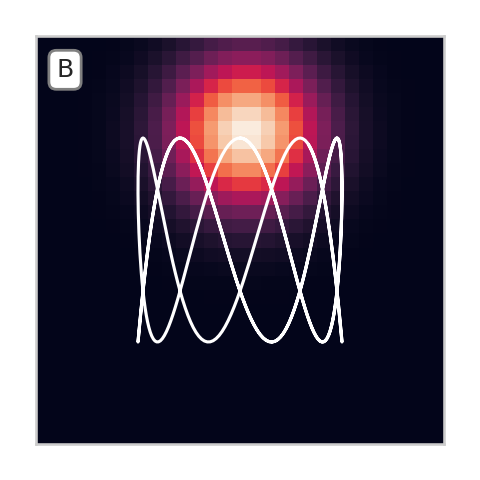}
  \end{minipage}
  \caption{(A) Prediction error of Lissajous blob with $\alpha=0.3$, $\beta=1$
    The ESN predicts almost to machine precision.
  (B) Gaussian blob and the path of its maximum (white line).}
  \label{fig:lissajous_pred}
\end{figure}

We now progress to predicting image sequences, i.e.~time-series with hundreds
of spatially correlated variables. Specifically, the input frames we use have
a size of $30\times30$ pixels.
We first train our ESN to predict video of Gaussian blobs that move along
Lissajous curves, i.e., the center of the Gaussian moves
according to
\begin{align}
  \label{eq:lissajous}
  x(t) &= \sin(\alpha t),\\
  y(t) &= \cos(\beta t).
\end{align}
To make it possible for the ESN to store a sufficiently
long history in its internal state, we use a network with a hidden state size of
10000.
Fig.~\ref{fig:lissajous_pred}A shows that our spatial ESN is able to learn
(almost arbitrarily) complicated periodic systems. For fully periodic systems
the cycle-based prediction by construction cannot be beat, because it
reconstructs the paths perfectly. However, the ESN also predicts the
trajectories nearly to machine precision without knowing the cycle lengths before-hand. The input map for this task is a combination of all the
available functions that we introduced in Sec.~\ref{sub:extending_the_esn}.
A table with all input map parameters that we use as parameters for the spatial
ESN throughout this paper is listed in Fig.~\ref{tab:input_maps}.
As before, the ESN was trained on $L_{train}=2000$ frames and the LSTM on sub-sequences
of length 200.
As a LSTM state size of 10000 might be too large, we also trained smaller networks,
but without significantly better performance compared to the other
prediction methods in Fig.~\ref{fig:lissajous_pred}A.
While it looks like the LSTM is as bad as the trivial prediction, it actually
achieves errors about half of the trivial method. This is of course still nowhere
near the machine precision predictions of the other methods.
Animations that compare ESN, cycle-based, and LSTM predictions can be found
at \cite{supplements}.
One optimization of the ESN output layer in this case takes roughly \textbf{1.5 minutes},
while training of an LSTM of the same size takes longer than \textbf{2700 minutes}
on an AMD Ryzen Threadripper 1950X (32 core CPU).

\begin{figure}
  \centering
  \begin{minipage}{.36\linewidth}
    \includegraphics[width=\linewidth]{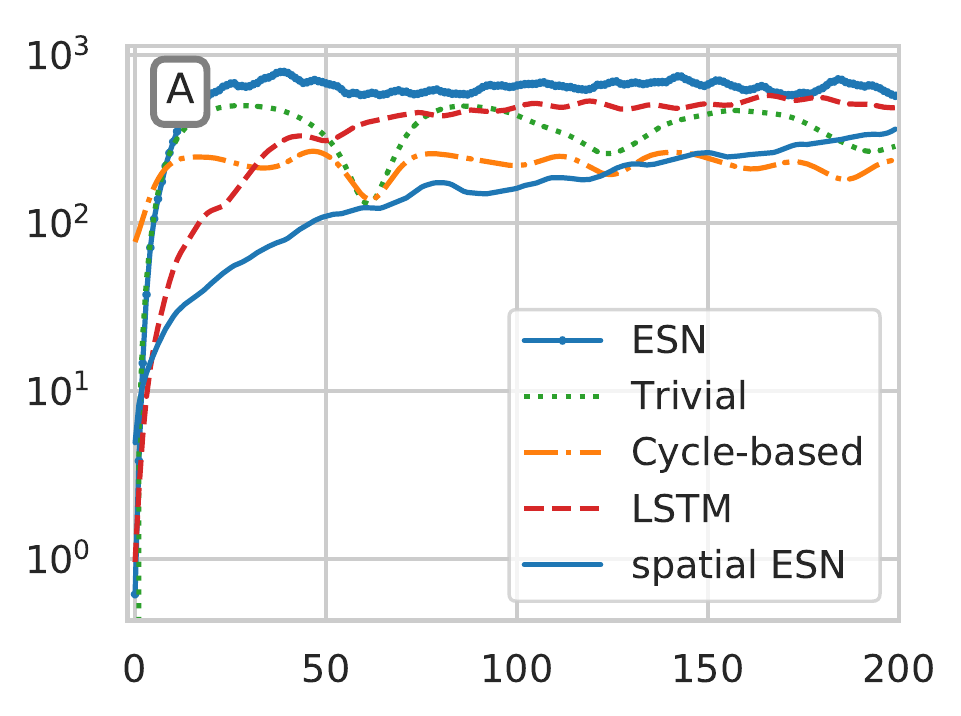}
  \end{minipage}
  \begin{minipage}{.265\linewidth}
    \includegraphics[width=\linewidth]{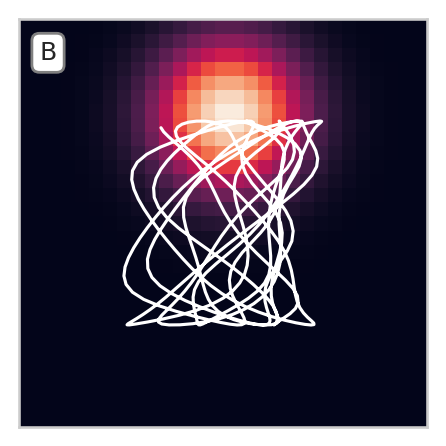}
  \end{minipage}
  \caption{(A) Prediction error on a chaotically moving blob in (B).
  The prediction of the basic ESN quickly deteriorates, while the LSTM
  produces good predictions for much longer.
  The spatial ESN beats the LSTM prediction almost by an order of magnitude.
  Supplementary material with animations of
  the predictions can be found in \cite{supplements}.}
  \label{fig:mackey_pred}
\end{figure}

\begin{figure}
  \centering
  \includegraphics[width=.7\linewidth]{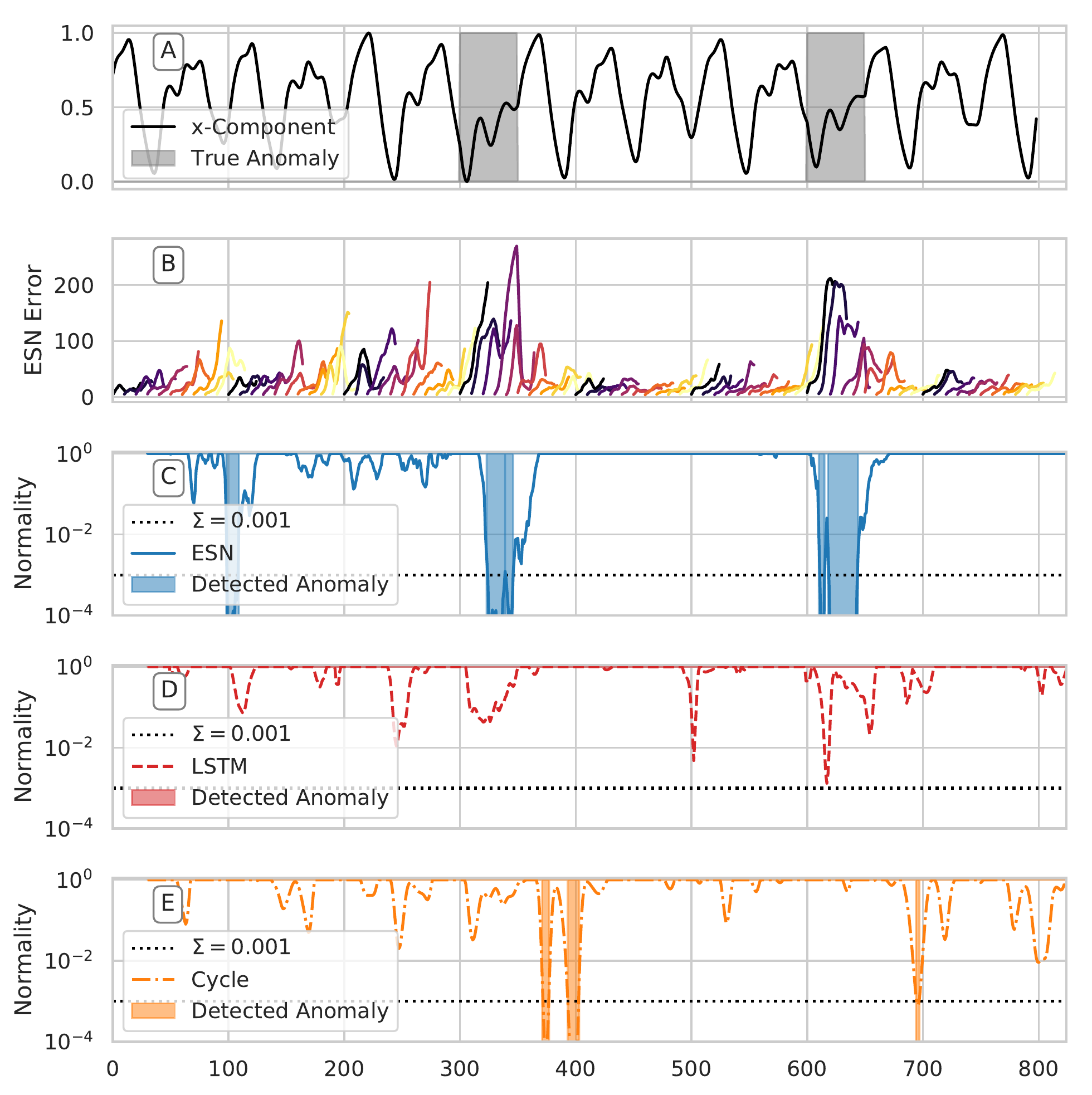}
  \caption{
    (A) The $x$-component of the Lissajous figure that produced the chaotically moving blob,
    with anomalous regions in gray.
    Plots (B), (C), and (D) show normality score for ESN, LSTM, and cycle-based predictions.
    Although difficult to detect for a human, our ESN clearly detects both anomalies
  (shaded regions) correctly. The LSTM does not find any anomalous behaviour and
  cycle based prediction detects an anomalies at the wrong times.
  }
  \label{fig:res_mackey_detect}
\end{figure}

Next we create a chaotic Lissajous figure by replacing $x(t)$ with the Mackey Glass time
series. The resulting prediction performance can be seen in
Fig.~\ref{fig:mackey_pred}A. The basic ESN is not able to reliably predict the chaotic
time series. The LSTM is again outperformed by our spatial ESN.
Next, we introduce anomalies in the MG time series, just as before
in the 1D case.\footnote{To a human observer they are practically invisible, as in both cases the
blob seems to move randomly.}  The spatial ESN detects both
anomalies, as seen in Fig.~\ref{fig:res_mackey_detect}.
The networks were trained on $L_{train} = 2000$ frames and we use $L_{pred} = 25$
for the anomaly detection. All other hyper-parameters remain the same.

\subsection{Kuroshio}%
\label{sub:kuroshio}

\begin{figure}
  \centering
  \includegraphics[width=0.7\linewidth]{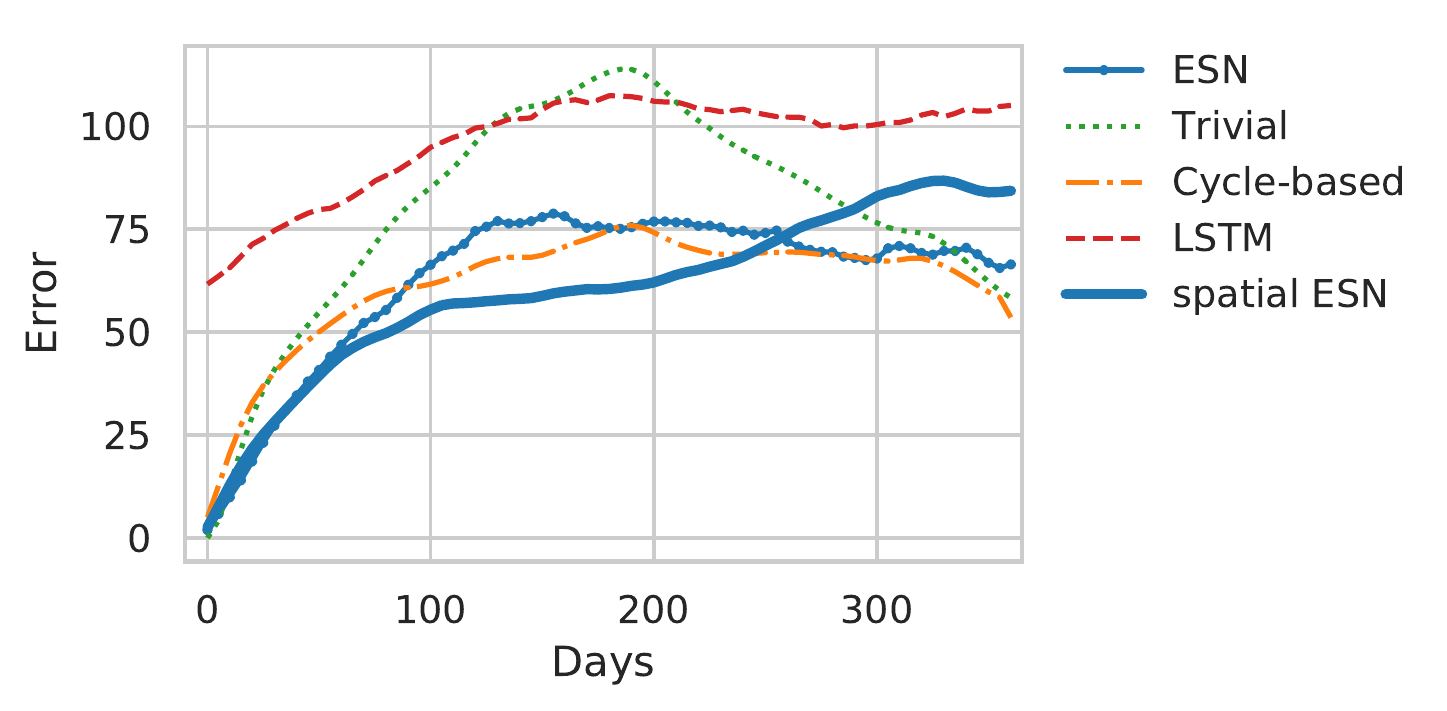}
  \caption{Averaged prediction performance over 100 different iterations.
  Under the IMED norm, spatial ESN predictions are quantitatively only slightly better than cycle-based
  and basic ESN, but qualitatively much more realistic
  (visible in \cite{supplements}).}
  \label{fig:res_pred_perf_kuro}
\end{figure}

The Kuroshio time series consists of 6435 days (17.6 years)
in steps of 3-day SSH means, resampled to 1287 5-day steps to make the length of a year an integer.
We let $L_{trans}=146$ (two years) and
$L_{train} = 730$ (10 years).  The averaged performance over 100 iterations is shown
in Fig.~\ref{fig:res_pred_perf_kuro}.
The LSTM converged to predicting the mean of the training sequence, which is a
common problem of RNNs.
The spatial ESN is better than the cycle-based predictions and the basic ESN
when predicting up to around 200 days ahead, but becomes worse after that.
The animations in the
supplementary material \cite{supplements}
show how the cyclic predictions repeat the same periodic fluctuations on top of the starting frame,
while ESN predictions are much more
 dynamic and actually look like possible continuations of the systems evolution.
For the anomaly detection we use $L_{pred} = 36$ (about half a year).

\begin{figure}
  \begin{minipage}{.63\linewidth}
    \includegraphics[width=\linewidth]{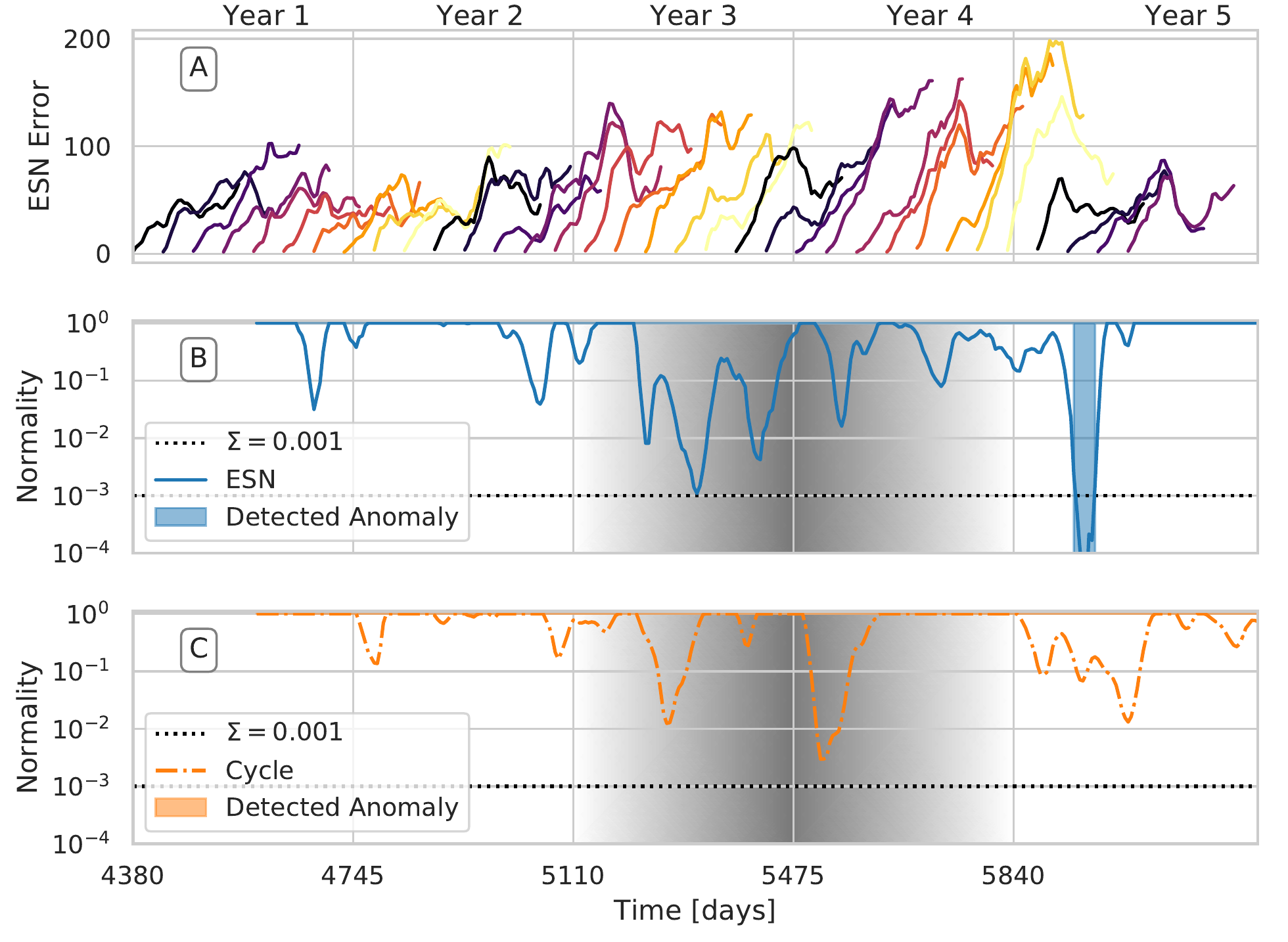}
    \caption{
      (A) Absolute IMED errors for ESN half-year predictions on ocean data.
      Plots (B), (C), and (D) show normality score for ESN, LSTM, and cycle-based predictions.
      The $x$-axis shows absolute simulation time.      
      The Kuroshio anomaly should be visible in the 3$^{\text{rd}}$ and
      4$^{\text{th}}$ predicted year (shaded in grey).
      The ESN prediction error grows during the anomaly, but not fast enough to
      trigger the threshold: The anomaly is localized, and its signal is obscured
      by the larger surroundings behaving normally, indicating
      that a spatially resolved error is needed
    }
    \label{fig:res_kuro_detect}
  \end{minipage}
  \hspace{0.01\linewidth}
  \begin{minipage}{.35\linewidth}
    \rowcolors{2}{gray!25}{white}
    \begin{tabular}{| l | r | r |}
      \hline \rowcolor{gray!50}
      Type & Size & Scale \\ \hline \hline
      Pixels & 30x30 & 3.0 \\ \hline
      Gauss. Conv. & 5x5 & 2.0 \\ \hline
      Gauss. Conv. & 10x10 & 1.5 \\ \hline
      Gauss. Conv. & 15x15 & 1.0 \\ \hline
      Random Conv. & 5x5 & 1.0 \\ \hline
      Random Conv. & 10x10 & 1.0 \\ \hline
      Random Conv. & 20x20 & 1.0 \\ \hline
      DCT & 15x15 & 1.0 \\ \hline
      DCT & 15x15 & 1.0 \\ \hline
      Gradient & 30x30 & 1.0 \\ \hline
      Gradient & 30x30 & 1.0 \\ \hline
      \end{tabular}
    \caption{Input maps of the spatial ESN used during this paper. The input
    scale parameters are set such that the values of each function take on
    approximately the same range of values.}
    \label{tab:input_maps}
  \end{minipage}
\end{figure}

Running the anomaly detection over the whole data set of 5-day averages
results in Fig.~\ref{fig:res_kuro_detect}.
The real Kuroshio anomaly starts somewhere around Day 5200 (3$^{\text{rd}}$
predicted year) and continues until the end.
Both Fig.~\ref{fig:res_kuro_detect}B and C show a clear signal of decreased normality
during the anomaly, but not sharp enough to trigger the normality score threshold.
The reason for the inconclusive normality score
sequences is that our error metric averages over a whole frame.
The anomaly is localized to a smaller region, which means that the well-predicted
areas away from it dillute the error-signal. This indicates that we need to resolve
our error measurement down into smaller spatial regions. Many schemes are possible,
and we discuss some of them in Section \ref{sec:conclusion}.

\begin{figure}
  \centering
  \includegraphics[width=\linewidth]{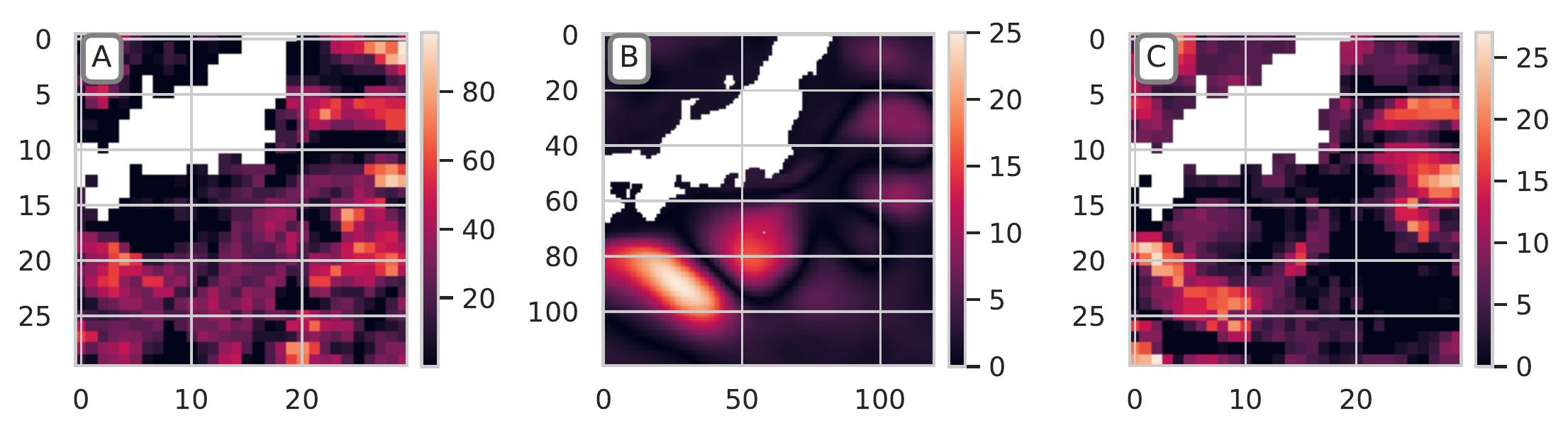}
  \caption{Anomaly count maps of (A) cycle-based, and (C) ESN prediction.
    (B) shows Fig.~2C as reference.
  Each cell contains the number of times a given pixel was anomalous over
  the whole time series. The Kuroshio anomaly is only clearly visible in the ESN
  prediction.}
  \label{fig:res_kuro_anomaly_count}
\end{figure}

For the present work, we simply look at the errors of individual
grid cells over time, compute element-wise normality scores and threshold them with the usual
$\Sigma=0.001$. After summing up all instances of $\Sigma<0.001$ we are left with a
map in which each cell represents the number of anomalies at that pixel
over the whole time series.
These maps are shown in
Fig.~\ref{fig:res_kuro_anomaly_count} for all three methods and indicate where
in space anomalies occur frequently.
Plot (A) shows the anomaly count resulting from the cycle-based,
and (C) the ESN prediction. (B) shows Fig.~2C, which we take as a reference ``snapshot''
of the anomaly.

Plot (C) shows a large region with high anomaly counts in the bottom left and a less
anomalous patch in the turbulent regions on the right.
Comparing Fig.~\ref{fig:res_kuro_anomaly_count}C to Fig.~\ref{fig:intro_kuroshio_elon_contr}C, we see
that our ESN has successfully detected all the main features of the Kuroshio anomaly.
The anomaly counts from the cycle-based predictions in (A)
do not reveal the true anomaly features,
but only show false positives in the more turbulent parts of the region.

\begin{figure}
  \centering
  \includegraphics[width=.8\linewidth]{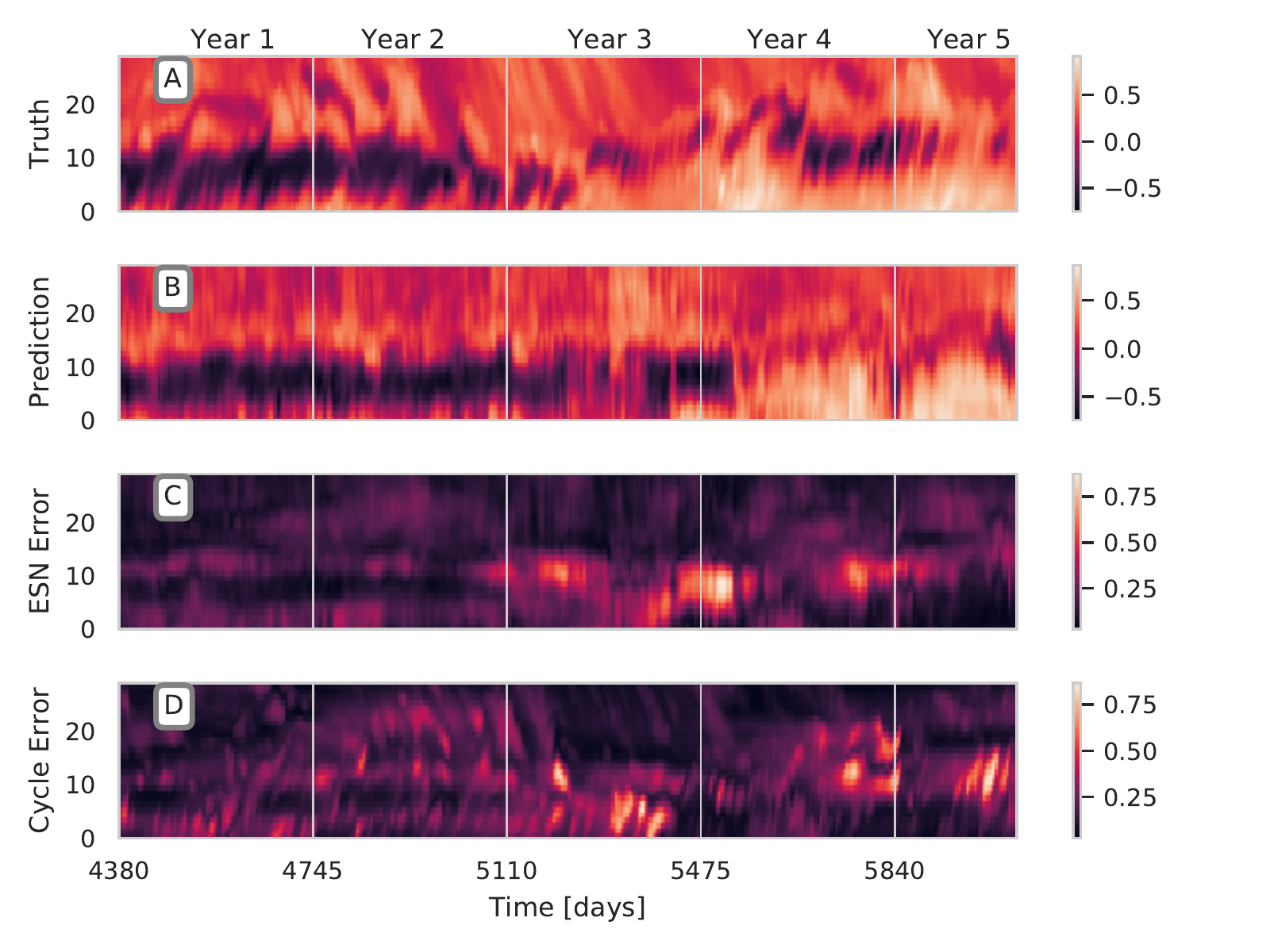}
  \caption{
    (A) True evolution of 25th row of the ocean simulation over time -
    The Kuroshio anomaly is nicely visible around year 3.
    (B) Row as predicted by the ESN half a year before.
    (C) ESN prediction error sequence.
    (D) Cycle-based prediction error.
    Both detection techniques show increased errors from around year 3,
    but the ESN prediction is much more accurate, reducing risk
    for false positives. The $x$-axis shows absolute simulation time.
  }
  \label{fig:res_kuro_detect_row}
\end{figure}

The anomaly count map allows us to automatically discover \emph{where} to look,
providing regions where we are likely to find anomalies.
To locate the Kuroshio anomaly in time we can examine a column of the input frames
that lies in the region with high anomaly count.
Fig.~\ref{fig:res_kuro_detect_row}A shows the true evolution of column 5 over time.
Plot B and C show ESN prediction and error respectively, which clearly indicates
an anomaly from the 3$^{\text{rd}}$ prediction year (around day 5475), where it actually is.

\section{Discussion and Future Work}%
\label{sec:conclusion}

With the simple means of the spatial ESNs described in
this paper, it was possible to predict spatio-temporal time series,
including turbulent ocean surface height simulations comprising
900 variables per time step, with surprising
precision: well enough that true anomalies could be detected whenever
our predictions failed, without being overwhelmed with false
positives.

Having successfully detected the Kuroshio anomaly with automatic methods,
the next stage for this work is to scale up the methods to discover
new ocean physics ``in the wild'', i.e., to search the full ocean
simulation data for unknown anomalies. While the per-pixel anomaly count
was sufficient to localize the Kuroshio without false positives, we
expect that it would yield more false positives in areas with higher
turbulence, as even near-perfect predictions would not yield
pixel-correct predictions due to chaoticity, but ``similar'' turbulence patterns
displaced in space and time.
The automatic spatial localization mechanism can be made more resilient
to turbulence in many ways: a combination of localization and de-localization
can be realized by e.g.~computing $\Sigma_t$ on a wavelet basis instead of
per-pixel. This would also make it possible to build a hierarchical error measure,
letting us ``zoom in'' from larger areas to small ones, according to the calculated
likelihood of them containing an anomaly.
One can additionally include errors for other properties than cell values: e.g.
velocities, momenta, frequencies, field curl, and so on.
As well, extending the IMED to include smoothing over
time would make errors more robust to feature displacement in both space and time; this can be
done efficiently due to the separability of the kernel.

Our present work used only sea surface height information, but the full datasets
include also pressure, temperature, density, and many other physical properties that
cross-correlate with each other, and together can improve prediction. Handling
multiple image series does not require new theory, but does need some technical work.

Finally, while the computations shown in the present paper can be
performed on a laptop with timings measured in minutes, the large
scale problem of analysing multi-property full-world simulation data requires
improvements in efficiency. We are in the process of porting Torsk from pure
NumPy to NumPy+Bohrium \cite{kristensen2014, kristensen2016b, kristensen2016a} for
automatic deployment on GPU and massively parallel systems,
yielding both orders of magnitude faster runtimes and scalability to huge system sizes
through automatic streaming.

The search for unknown ocean phenomena will be carried out in collaboration with Team
Ocean at University of Copenhagen, who has identified six world regions, where
the likelihood of modal ocean currents existing is high.

While our present work focuses on oceanographic
simulation data, the methods are very general and can be applied to a
wide range of problems. We invite the reader to do so using our open source
implementation at \url{https://github.com/nmheim/torsk}.

\section{Acknowledgments}
\label{sec:acknowledgments}
James Avery was funded by the VILLUM Foundation (Villum Experiment
Project 00023321, ``Folding Carbon: A Calculus of Molecular
Origami'').\\
Niklas Heim was funded by the Czech Science Foundation (grants no.18-21409S)
and the OP VVV MEYS project CZ$.02.1.01/0.0/0.0/16$\_$019/0000765$ ``Research Center for Informatics''.

\bibliography{main}
\bibliographystyle{unsrtnat}
\appendix
\section{Bifurcations in RNN State Space}
\label{sec:bifurcations}

\begin{figure}
  \centering
  \includegraphics[width=.8\linewidth]{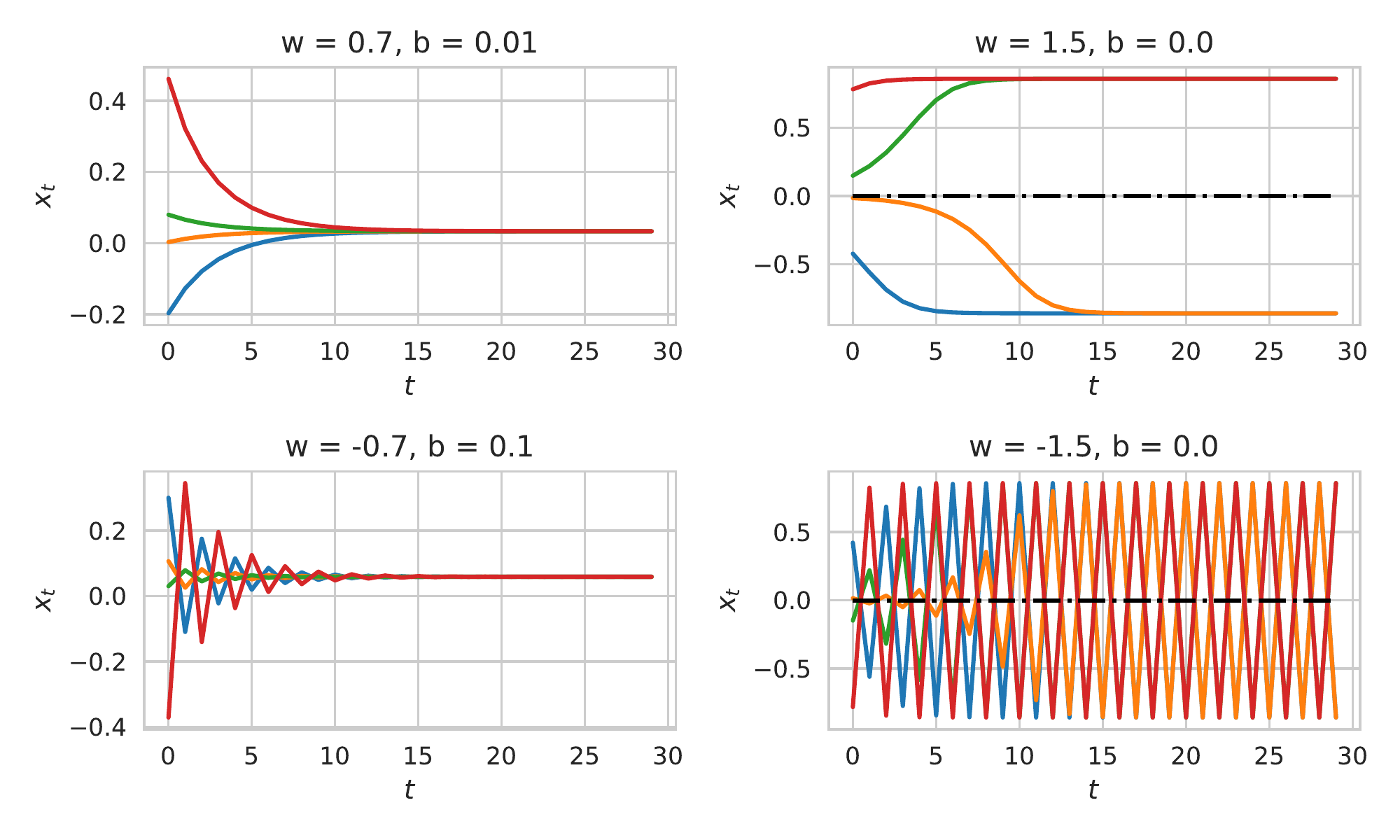}
  \caption{Evolution of $x_t$ over time for different parameters $w$ and $b$.
    Dashed lines show unstable fixed points. Apart from the expected fixed points
    that $x_t$ converges to over time, there are also oscillations visible in
    the last plot. Such oscillations that repeat every 2 iterations are called
    \emph{period-2 cycles} and they appear when $w<-1$.}
  \label{fig:bif_evolution}
\end{figure}

A problem that arises with the optimization of recurrent weights is that
the state space is not necessarily continuous, which was shown by \cite{doya1993}.
The points at which the state space can have discontinuities are
called \emph{bifurcations} and they can impair the learning
or prevent convergence to a local minimum completely. To understand what
bifurcations are and how they affect RNN training, we consider the
recurrent part of a single unit RNN with the hyperbolic tangent as the
activation function.  If the RNN has only one unit, the state $\vt{x}$, weights and biases become a scalars:
\begin{equation}
  \label{eq:single_unit_rnn}
  x_{t+1} = \tanh(w x_t + b).
\end{equation}
The parameter $w$ denotes the scalar weight of the single unit and $b = w_{in}
u_t$ will serve as the bias of a constant input of $u_t = 1$.  In
Fig.~\ref{fig:bif_evolution} we can see the evolution of $x_t$.  Depending on
different initial values $x_0$ and network parameters, the state converges to
different values for $t$ towards infinity. These values are called \emph{fixed
points} $x^*$ and for them $x_t = x_{t+1}$ holds.  In particular, fixed points
that the state converges to are called \emph{stable} fixed points (or
\emph{attractors}).  The second kind of fixed points are \emph{unstable}.  The
slightest deviation from an unstable fixed point will result in a flow away
from the point, which is why they are also called \emph{repellers}.  In the
first three cases of Fig.~\ref{fig:bif_evolution} a fixed point is always
reached. The fourth example in the lower right shows representatives of
the oscillating fixed point, more specifically \emph{period-2 cycles}, that
repeat every second iteration.
\begin{figure}
  \centering
  \includegraphics[width=.8\linewidth]{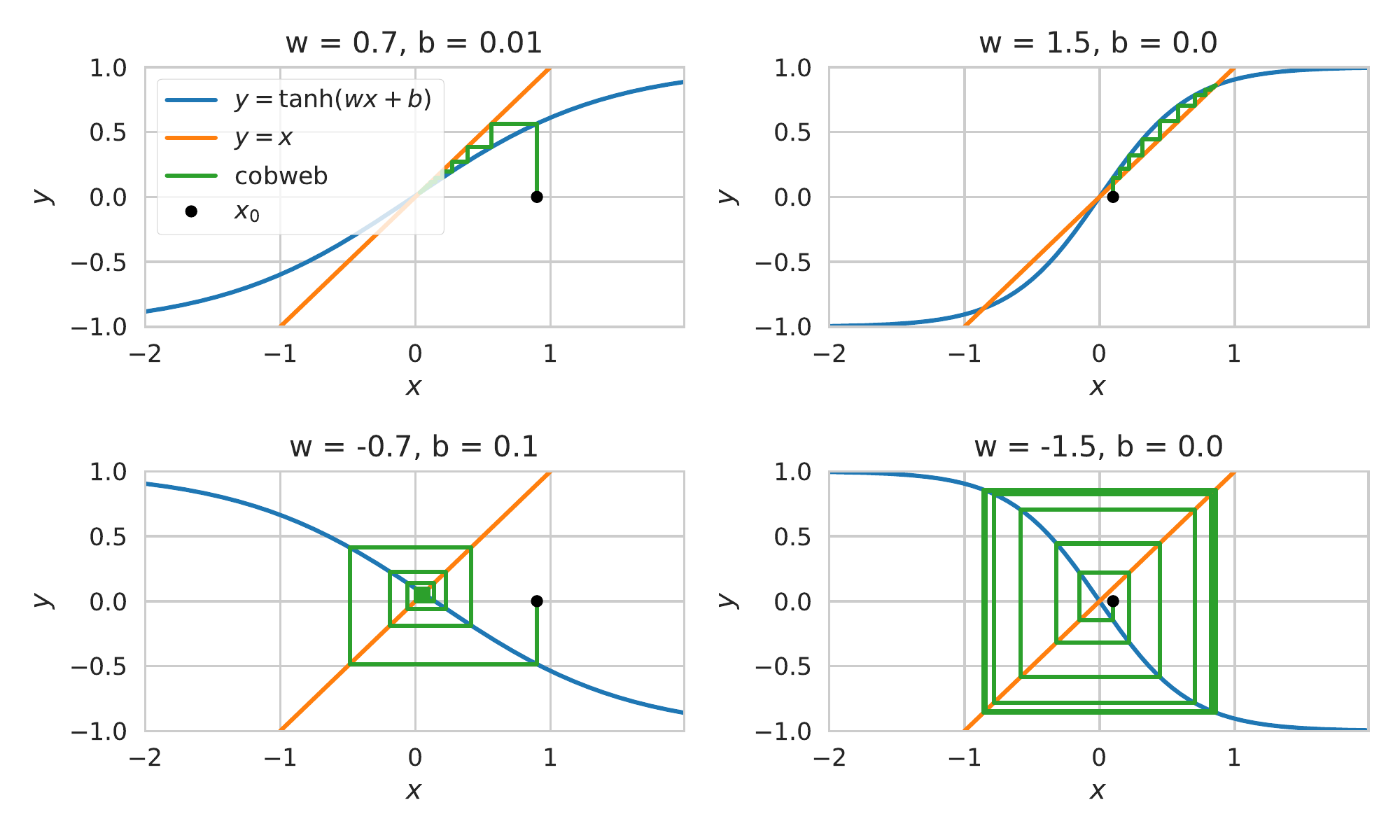}
  \caption{Cobwebs for the same parameters as in Fig.~\ref{fig:bif_evolution}.
    The black dot is the initial value $x_0$. By drawing a vertical line to the
    intersection with the activation function gives the new input $x_1$. Drawing
    a horizontal line to the intersection with $y = x$ projects the point back to
    the $x$-axis.  The projection is the next input to the activation function.
    This process is repeated until a stable orbit or a fixed point is reached.}
  \label{fig:cobweb}
\end{figure}
By varying the parameters $w$ and $b$ the location and the nature of fixed
points can be changed. The blue line in the right plot of
Fig.~\ref{fig:fixed_points} splits in two as $w$ is increased. The point at
$w=1$ is called a \emph{bifurcation} point.  There are two things that are
happening here: the stable fixed point at $x=0$ becomes unstable (indicated by
the dashed line) and two new stable fixed points above and below zero are
created. For an in depth introduction to chaotic systems we refer to \cite{strogatz}.

A mathematical analysis of fixed points can be done by assuming that $x^*$ is a
fixed point we can analyze Eq.~\eqref{eq:single_unit_rnn}:
\begin{equation}
  \label{eq:fp}
  x^* = \tanh(wx^* +b).
\end{equation}
Solving once for $w$ and once for $b$ results in two equations for fixed
points:
\begin{align}
  b &= \tanh^{-1}(x) - wx\\
  w &= \frac{\tanh^{-1}(x) - b}{x},
\end{align}
which can be plotted for different values of $w$ and $b$
(Fig.~\ref{fig:fixed_points}).  The period-2 cycles cannot be found by
analysing Eq.~\eqref{eq:fp}.  Instead they can be found analytically by solving
\begin{equation}
  x^* = \tanh^2(wx^* + b),
\end{equation}
but also by an intuitive, graphical approach called \emph{cobwebbing} (Fig.~\ref{fig:cobweb}).
Starting from an initial point $x_0$ a vertical line is drawn to the value of
the activation function. Now drawing a horizontal line until we intersect with
the graph of $y = x$ gives the new input $x_1$ and so forth.\\
Now that we have an understanding of what fixed points and bifurcations are we
can examine their effect on RNN learning. Suppose we initialize the network
with a constant $b=0.1$ and a $w=3$. If $x_0$ is negative, the nearest fixed
point is on the lower branch of the yellow line in the right plot of
Fig.~\ref{fig:fixed_points}.  Further assume we train the network to output
$x_\infty = - 0.25$.  In this case, $w$ will be lowered to approach $x^*= -
0.25$ until the bifurcation point is reached and the stable fixed point
vanishes (yellow line becomes dashed line). The fixed point becomes unstable
and the network output will change discontinuously as it jumps to the attractor
on the upper branch.  This will result in a discontinuity in the loss function
and an infinite gradient.  After jumping to the upper branch $w$ will grow
towards infinite values as the GD algorithm tries to approach the target value
of $x = - 0.25$.  Similar examples can be constructed in which parameters
oscillate between two bifurcation points.\\
The weights of RNNs are normally initialized to very small values which results
in few fixed points. As the network learns some of the weights increase which
drives the RNN through bifurcations.  The discontinuities that result in very
large gradients cause large jumps of the GD algorithm which can nullify the
learning of hundreds of steps in a single iteration. Aside from the vanishing
and exploding gradient problems, bifurcations are another major reason for the
intricacy of RNN training.
\begin{figure}
  \centering
  \includegraphics[width=.8\linewidth]{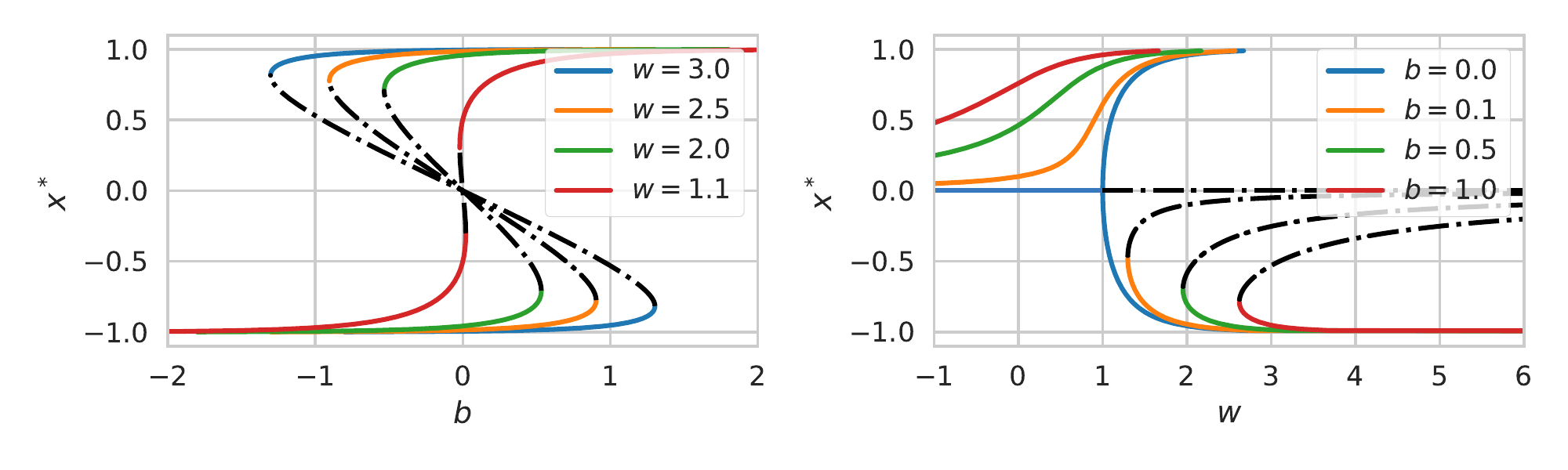}
  \caption{Fixed points for different parameter values of $b$ and $w$.  The
    values of the fixed points $x^*$ are affected by varying the weights of the
    RNN.  If there is more than one stable fixed point $x_t$ converges to the
    attractor that is closest to the initial value $x_0$.  Dashed lines denote
    unstable fixed points, which can only be reached if $x_0 = x^*$.}
  \label{fig:fixed_points}
\end{figure}
Keeping the weights fixed eliminates all three of theses problems. Although
ESN do not have the same expressiveness as a general RNN, it is computationally
strong enough to capture very complex behaviour.

\end{document}